%% file: icml2022_conference.tex
\documentclass[nohyperref]{article}

\usepackage{microtype}
\usepackage{graphicx}
\usepackage{booktabs} 

\usepackage{hyperref}

\usepackage[accepted]{arxiv}

\usepackage{amsmath}
\usepackage{amssymb}
\usepackage{mathtools}
\usepackage{amsthm}

\usepackage[capitalize,noabbrev]{cleveref}

\theoremstyle{plain}

\theoremstyle{definition}

\theoremstyle{remark}

\usepackage[textsize=tiny]{todonotes}

\usepackage{url}
\usepackage{graphicx}
\usepackage{subcaption}
\usepackage{booktabs}
\usepackage{multirow}
\usepackage{mathtools}
\usepackage{wrapfig}
\usepackage{enumitem}

\usepackage{bbm}
\usepackage{relsize}
\usepackage{lscape}

\input{math_commands.tex}

\icmltitlerunning{Neural Simulated Annealing}

\begin{document}

\twocolumn[
\icmltitle{Neural Simulated Annealing}



\icmlsetsymbol{equal}{*}

\begin{icmlauthorlist}
\icmlauthor{Alvaro H.C. Correia}{equal,yyy}
\icmlauthor{Daniel E. Worrall}{equal,comp}
\icmlauthor{Roberto Bondesan}{comp}
\end{icmlauthorlist}

\icmlaffiliation{yyy}{Eindhoven University of Technology, Eindhovem, the Netherlands (Work done during internship at Qualcomm AI Research)}
\icmlaffiliation{comp}{Qualcomm AI Research, Qualcomm Technologies Netherlands B.V. (Qualcomm AI Research is an initiative of Qualcomm Technologies, Inc.)}

\icmlcorrespondingauthor{Alvaro H.C. Correia}{a.h.chaim.correia@tue.nl}

\icmlkeywords{Machine Learning, ICML}

\vskip 0.3in
]



\printAffiliationsAndNotice{\icmlEqualContribution} 

\begin{abstract}
Simulated annealing (SA) is a stochastic global optimisation technique applicable to a wide range of discrete and continuous variable problems. Despite its simplicity, the development of an effective SA optimiser for a given problem hinges on a handful of carefully handpicked components; namely, neighbour proposal distribution and temperature annealing schedule. In this work, we view SA from a reinforcement learning perspective and frame the proposal distribution as a policy, which can be optimised for higher solution quality given a fixed computational budget. We demonstrate that this Neural SA with such a learnt proposal distribution, parametrised by small equivariant neural networks, outperforms SA baselines on a number of problems: Rosenbrock's function, the Knapsack problem, the Bin Packing problem, and the Travelling Salesperson problem. We also show that Neural SA scales well to large problems---generalising to significantly larger problems than the ones seen during training---while achieving comparable performance to popular off-the-shelf solvers and other machine learning methods in terms of solution quality and wall-clock time.
\end{abstract}

\section{Introduction}
There are many different kinds of combinatorial optimisation (CO) problem, spanning bin packing, routing, assignment, scheduling, constraint satisfaction, and more. Solving these problems while sidestepping their inherent computational intractability has great importance and impact for the real world, where poor bin packing or routing lead to wasted profit or excess greenhouse emissions \citep{salimifard2012green}. General solving frameworks or \emph{metaheuristics} for all these problems are desirable, due to their conceptual simplicity and ease-of-deployment, but require manual tailoring to each individual problem. 
%
One such metaheuristic is \emph{Simulated Annealing} (SA) \citep{kirkpatrick1987optimization}, a simple, and equally very popular, iterative global optimisation technique for numerically approximating the global minimum of both continuous- and discrete-variable problems. While SA has wide applicability, this is also its Achilles' Heel, leaving many design choices to the user. Namely, a user has to design 1) neighbourhood proposal distributions, which define the space of possible transitions from a solution $\rvx_k$ at time $k$ to solutions $\rvx_{k{+}1}$ at time $k{+}1$, and 2) a temperature schedule, which determines the balance of exploration to exploitation. In this work, we mitigate the need for extensive finetuning of SA's parameters by designing a learnable proposal distribution, which we show improves convergence speed with little computational overhead (limited to $\gO(N)$ per step for problem size $N$).


%
In recent years, research on approximate optimisation methods has been inundated by works in machine learning for CO (ML4CO) \citep{bengio2018}. 
A lot of the focus has been on end-to-end neural architectures \citep{bello16, vinyals2017pointer, dai2017, kool2018attention, emami2018, bresson2021transformer}. These work by brute force learning the instance to solution mapping---in CO these are sometimes referred to as \emph{construction heuristics}. Other works focus on learning good parameters for classic algorithms, whether they be parameters of the original algorithm \citep{kruber2017, bonami2018} or extra neural parameters introduced into the computational graph of classic algorithms \citep{gasse2019exact, gupta2020, kool2021deep, costa2020, wu2021learning, chen2019learning,fu2021generalize}. Our method, \emph{neural simulated annealing} (Neural SA) can be viewed as sitting firmly within this last category.

\begin{figure*}[!ht]
    \small
    \centering
    \includegraphics[width=\textwidth]{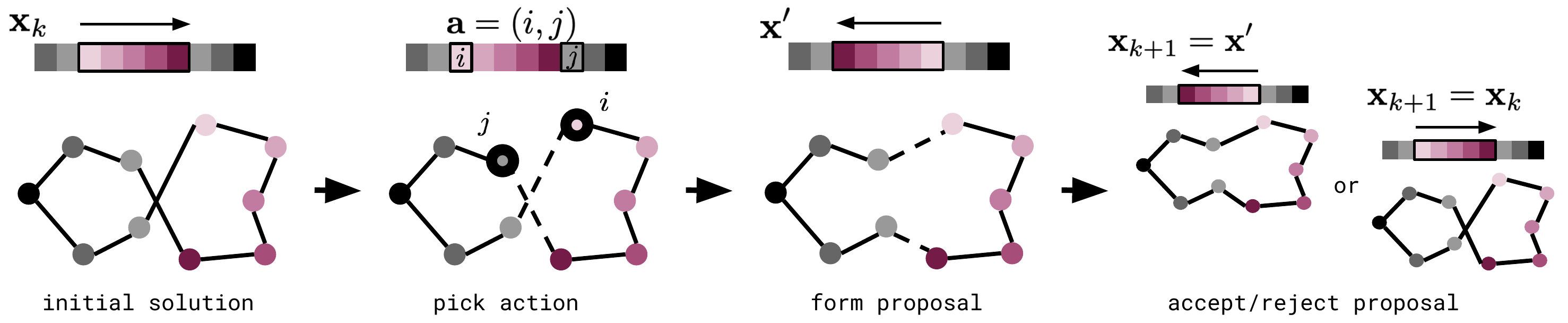}
    \caption{Neural SA pipeline for the TSP. Starting with a solution (tour) $\rvx_k$, we sample an action $\rva{=}(i,j)$ from our learnable policy/proposal distribution, defining start $i$ and end $j$ points of a 2-opt move (replacing two old with two new edges). Each pane shows both the linear and graph-based representations for a tour. From $\rvx_k$ and $\rva$ we form a proposal $\rvx'$ which is either accepted or rejected in a MH step. Accepted moves assign $\rvx_{k+1} {=} \rvx'$; whereas, rejected moves assign $\rvx_{k+1} {=} \rvx_k$.}
    \label{fig:neural_sa_pipeline}
\end{figure*}

SA is an \emph{improvement heuristic}; it navigates the search space of feasible solutions by iteratively applying (small) perturbations to previously found solutions. Figure~\ref{fig:neural_sa_pipeline} illustrates this for the Travelling Salesperson Problem (TSP), perhaps the most classic of NP-hard problems. 
In this work, we pose this as a Reinforcement Learning (RL) agent navigating an environment, searching for better solutions. In this light the proposal distribution is an optimisable quantity. Conveniently, our method inherits convergence guarantees from SA. We are able to directly optimise the proposal distribution using policy optimisation for both faster convergence and better solution quality under a fixed computation budget. We demonstrate Neural SA on four tasks: Rosenbrock's function, a toy 2D optimisation problem, where we can easily visualise and analyse what is being learnt; the Knapsack and Bin Packing problems, which are classic NP-hard resource allocation problems; and the TSP.

Our contributions are:
\begin{itemize}[itemsep=0pt,topsep=0pt]
    \item We pose simulated annealing as a Markov decision process, bringing it into the realm of reinforcement learning. This allows us to optimise the proposal distribution in a principled manner, still preserving all the convergence guarantees of vanilla simulated annealing.
    \item We show competitive performance to off-the-shelf CO tools and other ML4CO methods on the Knapsack, Bin Packing, and Travelling Salesperson problems, in terms of solution quality and wall-clock time.
    \item We show our methods transfer to problems of different sizes, and also perform well on problems up to $40\times$ larger than the ones used for training.
    \item Our method is competitive within the ML4CO space, using a very lightweight architecture, with number of learnable parameters of the order of 100s or fewer.
\end{itemize}

\section{Background and Related Work}
\label{sec:background}
Here we outline the basic simulated annealing algorithm and its main components. Then we provide an overview of prior works in the machine learning literature which have sought to learn parts of the algorithm or where SA has found uses in machine learning.

\paragraph{Combinatorial optimisation}
A combinatorial optimisation problem is defined by a triple $(\bm{\Psi}, \gX, E)$ where $\bm{\psi} \in \bm{\Psi}$ are problem instances (city locations in the TSP), $\gX$ is the set of feasible solutions given $\bm{\psi}$ (Hamiltonian cycles in the TSP) and $E: \gX \times \bm{\Psi} \to \sR$ is an \emph{energy function} (tour length in the TSP). Without loss of generality, the task is to minimise the energy $\min_{\rvx \in \gX} E(\rvx; \bm{\psi})$. CO problems are in general NP-hard, meaning that there is no known algorithm to solve them in time polynomial in the number of bits that represents a problem instance.

\paragraph{Simulated Annealing}
Simulated annealing \citep{kirkpatrick1987optimization} is a metaheuristic for CO problems. It builds an inhomogeneous Markov chain $\rvx_0 \to \rvx_1 \to \rvx_2 \to \cdots$ for $\rvx_k \in \mathcal{X}$, asymptotically converging to a minimizer of $E$. The stochastic transitions $\rvx_k \to \rvx_{k+1}$ depend on two quantities: 1) a proposal distribution, and 2) a temperature schedule. The proposal distribution $\pi: \mathcal{X} \to \sP(\mathcal{X})$, for $\sP(\mathcal{X})$ the space of probability distributions on $\gX$, suggests new states in the chain. It perturbs current solutions to new ones, potentially leading to lower energies immediately or later on. After perturbing a solution $\rvx_k \to \rvx'$, a Metropolis--Hastings (MH) step \citep{metropolis1953,hastings1970} is executed. This either accepts the perturbation ($\rvx_{k+1} = \rvx'$) or rejects it ($\rvx_{k+1} = \rvx_k$)---see Algorithm~\ref{alg:simulated-annealing} for details. The target distribution of the MH step has form $p(\rvx | T_k) \propto \exp \{ -E(\rvx) / T_k \}$, where $T_k$ is the \emph{temperature} at time $k$. In the limit $T_k \to 0$, this distribution tends to a sum of Dirac deltas on the minimisers of the energy. The temperature is annealed, according to the temperature schedule, $T_1, T_2, ...$, from high to low, to steer the target distribution smoothly from broad to peaked around the global optima. The algorithm is outlined in Algorithm~\ref{alg:simulated-annealing}. Under certain regularity conditions and provided the chain is long enough, it will visit the minimisers almost surely \citep{geman1984stochastic}. More concretely,
\begin{align}
    \lim_{k\to\infty} P \left ( \rvx_k \in \argmin_{\rvx \in \gX} E(\rvx; \bm{\psi}) \right) = 1. \label{eq:convergence}
\end{align}
Despite this guarantee, practical convergence speed is determined by $\pi$ and the temperature schedule, which are hard to fine-tune. There exist problem-specific heuristics for setting these \citep{pereira2004study,cicirello2007design}, but in this paper we propose to learn the proposal distribution.

\subsection{Simulated annealing and machine learning}
A natural way to combine machine learning and simulated annealing is to design local improvement heuristics that feed off each other. Cai et al.~\yrcite{cai2019reinforcement} and  Vashisht et al.~\yrcite{vashisht2020placement} use RL to find good initial solutions that are later refined by standard SA. That is fundamentally different to our approach, as we augment SA with RL-optimisable components, instead of simply using them as standalone algorithms that only interact via shared solutions. In fact, our method is perfectly compatible with theirs and any other SA application.
Another line of work seeks to optimise different components of SA with RL \citep{Wauters_2020,Khairy_2020,beloborodov2020reinforcement,mills2020finding} or statistical machine learning techniques \citep{blum2020learning}. In contrast to these methods that optimise individual hyperparameters in SA, we frame SA itself as an RL problem, which allows us to define and train the proposal distribution as a policy.

More closely to our method, other approaches improve the proposal distribution. 
In Adaptive Simulated Annealing (ASA) \citep{ingber1996adaptive} the proposal distribution is not fixed but evolves throughout the annealing process as a function of the variance of the quality of visited solutions. ASA improves the convergence of standard SA but is not learnable like Neural SA.
To the best of our knowledge, \cite{marcos2012supervised} are the only others to learn the proposal distribution for SA, but they rely on supervised learning, requiring high quality solutions or good search strategies to imitate; both expensive to compute. Conversely, Neural SA is fully unsupervised, thus easier to train and extend to different CO tasks. 
Finally, SA is also akin to Metropolis-Hastings, a popular choice for Markov Chain Monte Carlo (MCMC) sampling.
Noé et al.~\yrcite{noe2019boltzmann}, Albergo et al.~\yrcite{albergo2019flow} and de Haan et al.~\yrcite{dehaan2021scaling} recently studied how to learn a proposal distribution of an MCMC chain for sampling the Boltzmann distribution of a physical system. While their results serve as motivation for our methods, we investigate a completely different context and set of applications.


Lastly, our work falls under bi-level optimisation methods, where an outer optimisation loop finds the best parameters of an inner optimisation. This encompasses situations such as learning the parameters \citep{rasdi2015simulated} or hyperparameters of a neural network optimiser \citep{maclaurin2015gradientbased,andrychowicz2016learning} and meta-learning \citep{finn2017model}. However, most recent approaches assume differentiable losses on continuous state spaces \cite{likhosherstov2021debiasing,ji2021bilevel,vicol2021unbiased}, while we focus on the more challenging CO setting. We note, however, methods in \cite{vicol2021unbiased} are based on evolution strategies and could be used in the discrete setting.

\subsection{Markov Decision Processes}
Simulated annealing naturally fits into the Markov Decision Process (MDP) framework as we explain below.
An MDP $\gM = (\gS,\gA,R,P,\gamma)$ consists of \emph{states} $s \in \gS$, \emph{actions} $a \in \gA$, an \emph{immediate reward function} $R: \gS \times \gA \times \gS \to \sR$, a \emph{transition kernel} $P: \gS \times \gA \to \sP(\gS)$, and a \emph{discount factor} $\gamma \in [0, 1]$. On top of this MDP we add a stochastic \emph{policy} $\pi: \gS \to \sP(\gA)$. The policy and transition kernel together define a length-$K$ trajectory $\tau = (s_0, a_0, s_1, a_1, ..., s_K)$, which is a sample from the distribution $P(\tau | \pi) = \rho_0(s_0) \prod_{k=0}^{K-1} P(s_{k+1} | s_k, a_k) \pi(a_k | s_k)$ and where $s_0 \sim \rho_0$ is sampled from the \emph{start-state distribution} $\rho_0$. One can then define the \emph{discounted return} $R(\tau) = \sum_{k=0}^{K-1} \gamma^t r_{k}$ over a trajectory, where $r_k = R(s_k, a_k, s_{k+1})$. We say that we have solved an MDP if we have found a policy that maximises the expected return $\E_{\tau \sim P(\tau | \pi)}[R(\tau)]$.

\section{Method}
Here we outline our approach to learn the proposal distribution. First we define an MDP corresponding to SA. We then show how the proposal distribution can be optimised and provide a justification that this does not affect convergence guarantees of the classic algorithm.

\subsection{MDP Formulation}
We formalise SA as an MDP, with states $\rvs = (\rvx, \bm{\psi}, T) \in \gS$ for $\bm{\psi}$ a parametric description of the problem instance as in Section~\ref{sec:background}, and $T$ the instantaneous temperature. Examples are in Section~\ref{sec:experiments}. Our actions $\rva \in \mathcal{A}$ perturb $(\rvx, \bm{\psi}, T) \mapsto (\rvx', \bm{\psi}, T)$, where $\rvx' \in \gN(\rvx)$ is a solution in the neighbourhood of $\rvx$. It is common to define small neighbourhoods, to limit energy variation from one state to the next. This heuristic discards exceptionally good and exceptionally bad moves, but since the latter are more common than the former, it generally leads to faster convergence.

We view the MH step in SA as a stochastic transition kernel, governed by the current temperature of the system, with transition probabilities following a Gibbs distribution and dynamics
\begin{align}
    \nonumber \rvx_{k+1} &= \left \{ \begin{matrix*}[l]
        \rvx',      & \text{with probability } p \\
        \rvx_{k},   & \text{with probability } 1-p,
    \end{matrix*} \right .  \\ 
    \text{where } p &= \min \left \{ 1, 
    e^{
    -\frac{1}{T_k}(E(\rvx'; \bm{\psi}) - E(\rvx_k; \bm{\psi}))
    } 
    \right \}.
\end{align}
This defines a transition kernel $P(\rvs_{k+1} | \rvs_k, \rva)$, where we have $\rvs_{k+1}{=}(\rvx_{k+1}, \bm{\psi}, T)$. For rewards, we use either the immediate gain $E(\rvx_{k}; \bm{\psi}) - E(\rvx_{k+1}; \bm{\psi})$ or the primal reward $-\delta_{k=K-1}\min_{\rvx \in \rvx_{1:k}} E(\rvx; \bm{\psi})$. We explored training with two different methods: Proximal Policy Optimisation (PPO)~\citep{schulman2017proximal} and Evolution Strategies (ES) \cite{salimans2017evolution}. The immediate gain works best with PPO, where at each iteration of the rollout, the immediate gain gives fine-grained feedback on whether the previous action helped or not. The primal reward works best with ES because it is non-local, returning the minimum along an entire rollout $\tau$ at the very end. We explored using the acceptance count but found that this sometimes led to pathological behaviours. 
Similarly, we tried the primal integral~\citep{bertold2013}, which encourages finding a good solution fast, but found we could not get training dynamics to converge.

\begin{algorithm}[!htb]
\caption{Neural simulated annealing. To get back to vanilla SA, replace the parametrised proposal distribution $\pi_\theta$ with a uniform distribution $\pi$ over neighbourhoods $\gN(\bullet)$.}
\label{alg:simulated-annealing}
\begin{algorithmic}
\REQUIRE Initial state $\rvs_0 = (\rvx_0, \bm{\psi}, T_0)$, proposal distribution $\pi$, transition function $P$, temperature schedule $T_1 \geq T_2 \geq T_3 \geq ...$, energy function $E(\bullet; \bm{\psi})$
\FOR{$k=1:K$}
    \STATE $\rva \sim \pi_\theta(\rvs_k)$ \COMMENT{Sample action}
    \STATE $u \sim \text{Uniform}(u; 0,1)$ \COMMENT{Metropolis--Hastings step}
    \IF{$u < \exp \left \{ -(E(\rvx'; \bm{\psi}) - E(\rvx_k; \bm{\psi})) / T_k \right \}$} 
        \STATE $\rvs_{k+1} \gets (\rvx', \bm{\psi}, T_{k+1})$ \COMMENT{Accept}
    \ELSE
        \STATE $\rvs_{k+1} \gets (\rvx_k, \bm{\psi}, T_{k+1})$ \COMMENT{Reject}
    \ENDIF
\ENDFOR
\end{algorithmic}
\end{algorithm}

\subsection{Policy Network Architecture}
\label{sec:arch}
\begin{figure}[!htb]
    \small
    \centering
    \includegraphics[width=0.475\textwidth]{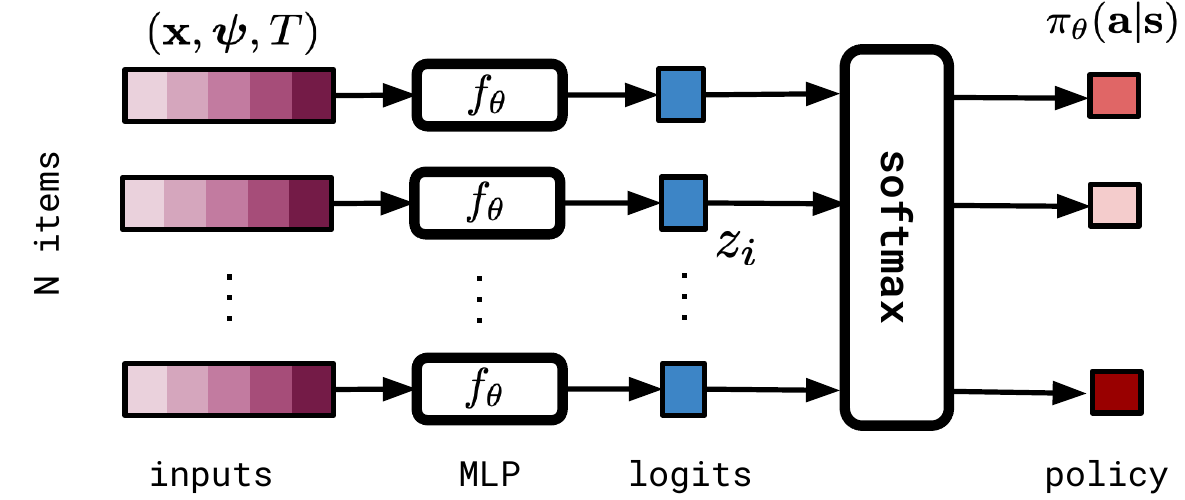}
    \caption{(a) Policy network used in all experiments. The same MLP is applied to all inputs pointwise.}
    \label{fig:policy-net}
\end{figure}

SA chains are long. It is because of this that we need as lightweight a policy architecture as possible. Furthermore, this architecture should have the capacity to scale to varying numbers of inputs, so that we can transfer experience across problems of different size $N$. We opt for a very simple network, shown in Figure~\ref{fig:policy-net}. For each dimension of the problem we map the state $(\rvx, \bm{\psi}, T)$ into a set of features. For all problems we try, there is a natural way to do this. Each feature is fed into an MLP, embedding it into a logit space, followed by a softmax function to yield probabilities. The complexity of this architecture scales linearly with $N$ and the computation is embarrassingly parallel,
which is important since we plan to evaluate it many times. A notable property of this architecture is that it is permutation equivariant~\citep{zaheer2017deep}---$\pi_\theta(\rva | \rvs) = \pi_\theta(\sigma \cdot \rva | \sigma \cdot \rvs)$ for $\sigma$ a permutation of $N$ objects---an important requirement for the CO problems we consider.
Note that our model is a permutation equivariant set-to-set mapping, but we have not used attention or other kinds of pairwise interaction to keep the computational complexity linear in the number of items. 

\paragraph{Convergence}
Convergence of SA to the optimum in the infinite time limit requires the Markov chain of the proposal distribution to be irreducible \cite{van1987simulated}, meaning that for any temperature, any two states are reachable through a sequence of transitions with positive conditional probability under $\pi$.
Our neural network policy satisfies this condition as long as the softmax layer does not assign zero probability to any state, a condition which is met in practice. Thus Neural SA inherits convergence guarantees from SA.

\section{Experiments} \label{sec:experiments}

\begin{figure*}[!htb]
    \small
    \begin{subfigure}[t]{0.24\textwidth}
    \vskip 7pt
         \centering
         \includegraphics[width=\textwidth]{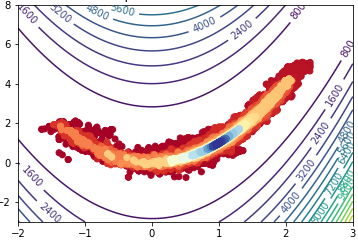}
         \caption{2D rollout}
         \label{fig:rb-basin}
     \end{subfigure}
     \hfill
     \begin{subfigure}[t]{0.24\textwidth}
     \vskip 0pt
         \centering
         \includegraphics[width=\textwidth]{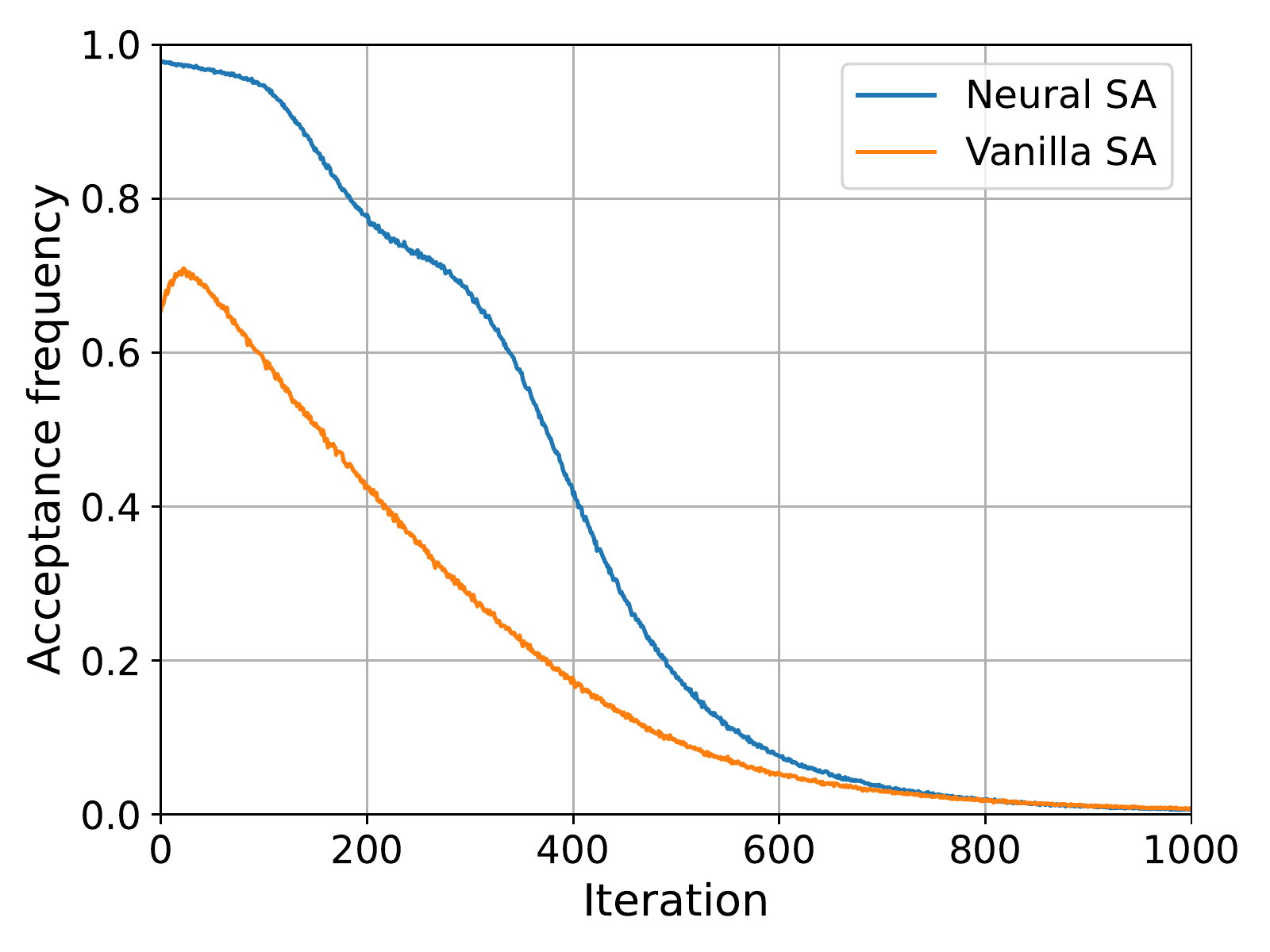}
         \caption{Acceptance ratios}
         \label{fig:rb-baseline-pacc}
     \end{subfigure}
     \hfill
     \begin{subfigure}[t]{0.24\textwidth}
     \vskip 0pt
         \centering
         \includegraphics[width=\textwidth]{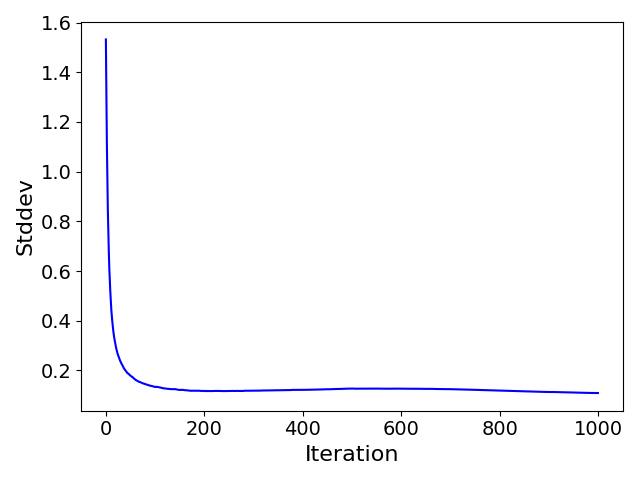}
         \caption{$\sigma$ by iteration}
         \label{fig:rb-es-std}
     \end{subfigure}
     \hfill
     \begin{subfigure}[t]{0.24\textwidth}
     \vskip 0pt
         \centering
         \includegraphics[width=\textwidth]{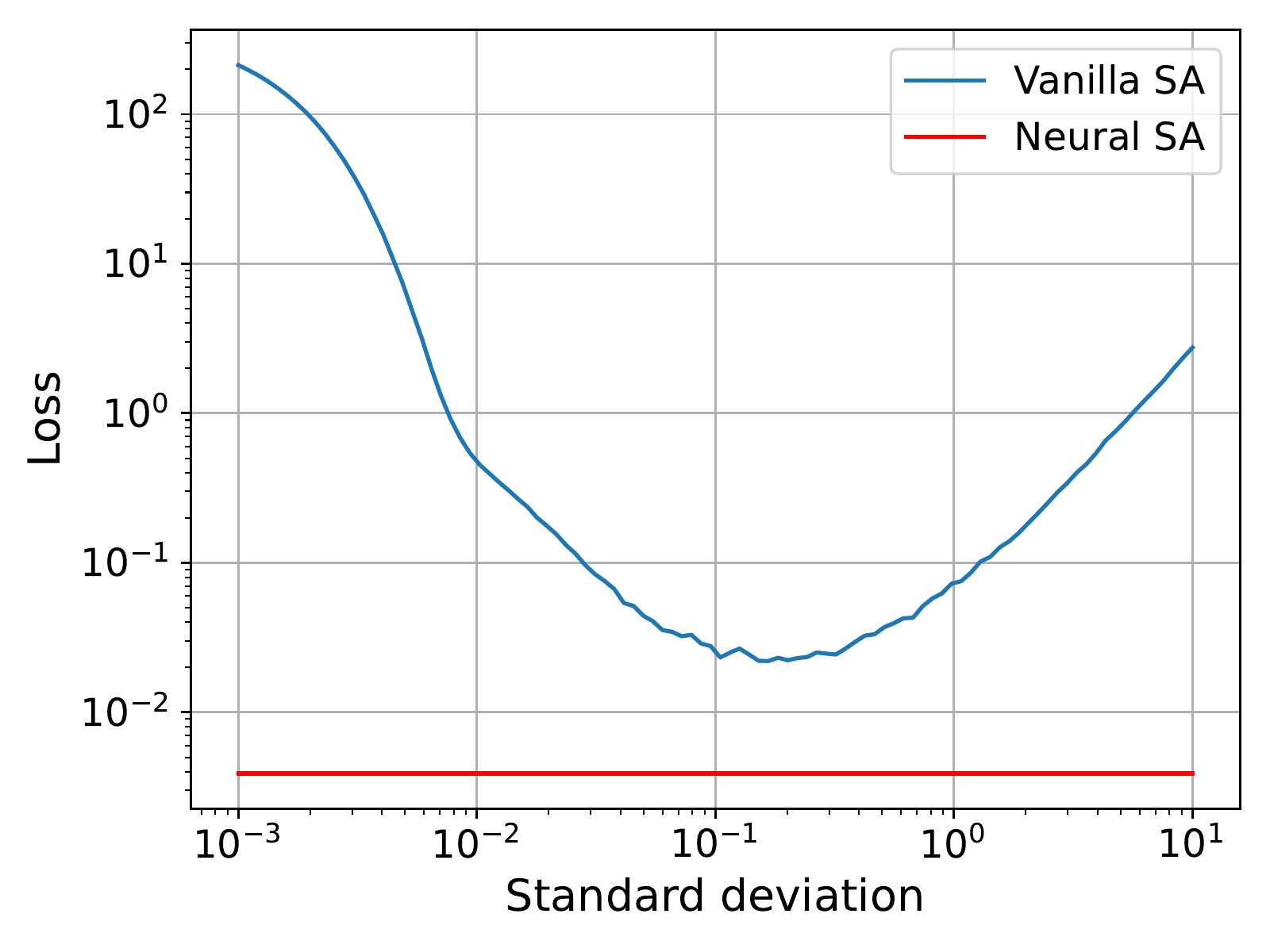}
         \caption{Baseline sweep}
         \label{fig:rb-baseline-sweep}
     \end{subfigure}
    \caption{Results on Rosenbrock's function: (a) Example trajectory, moving from red to blue, showing convergence around the minimiser at (1,1) (b) Neural SA has higher acceptance ratio than the baseline, a trend observed in all experiments, (c) Standard deviation of the learned policy as a function of iteration. Large initial steps offer great gains followed by small exploitative steps, (d) A non-adaptive vanilla SA baseline cannot match an adaptive one, no matter the standard deviation.
    }
    \label{fig:rb-ppo}
\end{figure*}

We evaluate our method on 4 tasks---Rosenbrock's function, the Knapsack, Bin Packing, and TSP problems---\emph{using the same architecture and hyperparameters of Neural SA for all tasks}. This shows the wide applicability and ease of use of our method. For each task (except for Rosenbrock's function) we test Neural SA on problems of different size $N$, \emph{training only on the smallest}. Similarly, we consider rollouts of different lengths, \emph{training only on short ones}. This accelerates training, showing Neural SA's generalisation capabilities. This type of transfer learning is one of the challenges in ML4CO \cite{joshi2019learning}, and is a merit of our lightweight, equivariant architecture. In all experiments, we start from trivial or random solutions and adopt an exponential multiplicative cooling schedule as originally proposed by \cite{kirkpatrick1987optimization}, with $T_k = \alpha^{k} T_0$. In practice, we define the temperature schedule by fixing $T_0$, $T_K$ and computing $\alpha$ according to the desired number of steps $K$. This allows us to vary the rollout length while maintaining the same range of temperatures for every run. We provide more precise experimental details in the appendix.

\subsection{The Rosenbrock function} \label{sec:rosenbrock}
The Rosenbrock function is a common benchmark for optimisation algorithms. It is a non-convex function over Euclidean space defined as 
\begin{align}
    E(x_0, x_1; a, b) = (a-x_0)^2 + b(x_1 - x_0^2)^2,
\end{align}
and with global minimum at $\rvx {=} (a,a^2)$.
Of course, gradient-based optimisers are more suited to this problem, but we use it as a toy example to showcase the properties of Neural SA. Our policy is an axis-aligned Gaussian $\pi_\theta(\rva | \rvs) = \gN(\rva ; \bm{0}, \sigma_\theta^2(\rvs_k))$, where we parametrise the variance $\sigma_\theta^2$ by an MLP of shape $2 {\to} 16 {\to} 2$ with a ReLU in the middle. Proposals are of the form $\rvx' {=} \rvx {+} \rva$, and the state is given by $\rvs_k = (\rvx_k, a, b, T_k)$. An example rollout is in Figure~\ref{fig:rb-basin}.

We contrast Neural SA against vanilla SA with fixed proposal distribution, i.e. $\sigma(\rvs_i) = \sigma$, for different $\sigma$ averaged over $2^{17}$ problem instances. This shows in Figure~\ref{fig:rb-baseline-sweep} that no constant variance policy can outperform an adaptive policy on this problem. Plots of acceptance ratio in Figure~\ref{fig:rb-baseline-pacc} show Neural SA has higher acceptance probability early in the rollout, a trend we observed in all experiments, suggesting its proposals are skewed towards lower energy solutions than standard SA. Figure~\ref{fig:rb-es-std} shows the variance network $\sigma_\theta^2(\rvs_i)$ as a function of time. It has learnt to make large steps until hitting the basin, whereupon large moves will be rejected with high probability, so variance must be reduced.


\subsection{Knapsack Problem} \label{sec:knapsack}
The Knapsack problem is a classic CO problem in resource allocation. Given a set of $N$ items, each of a different value $v_i > 0$ and weight $w_i > 0$, the goal is to find a subset that maximises the sum of values while respecting a maximum total weight of $W$.
This is the 0-1 Knapsack Problem, which is weakly NP-complete, has a search space of size $2^N$ and corresponding integer linear program
\begin{align}
    \nonumber &\text{minimise } E(\rvx; \bm{\psi}) = -\sum_{i=0}^{N-1} v_i x_i, \\ &\text{subject to } \sum_{i=0}^{N-1} w_i x_i \leq W, \qquad x_i \in \{0, 1\}.
\end{align}

\begin{figure}[!hb]
    \small
    \centering
    \includegraphics[width=\columnwidth]{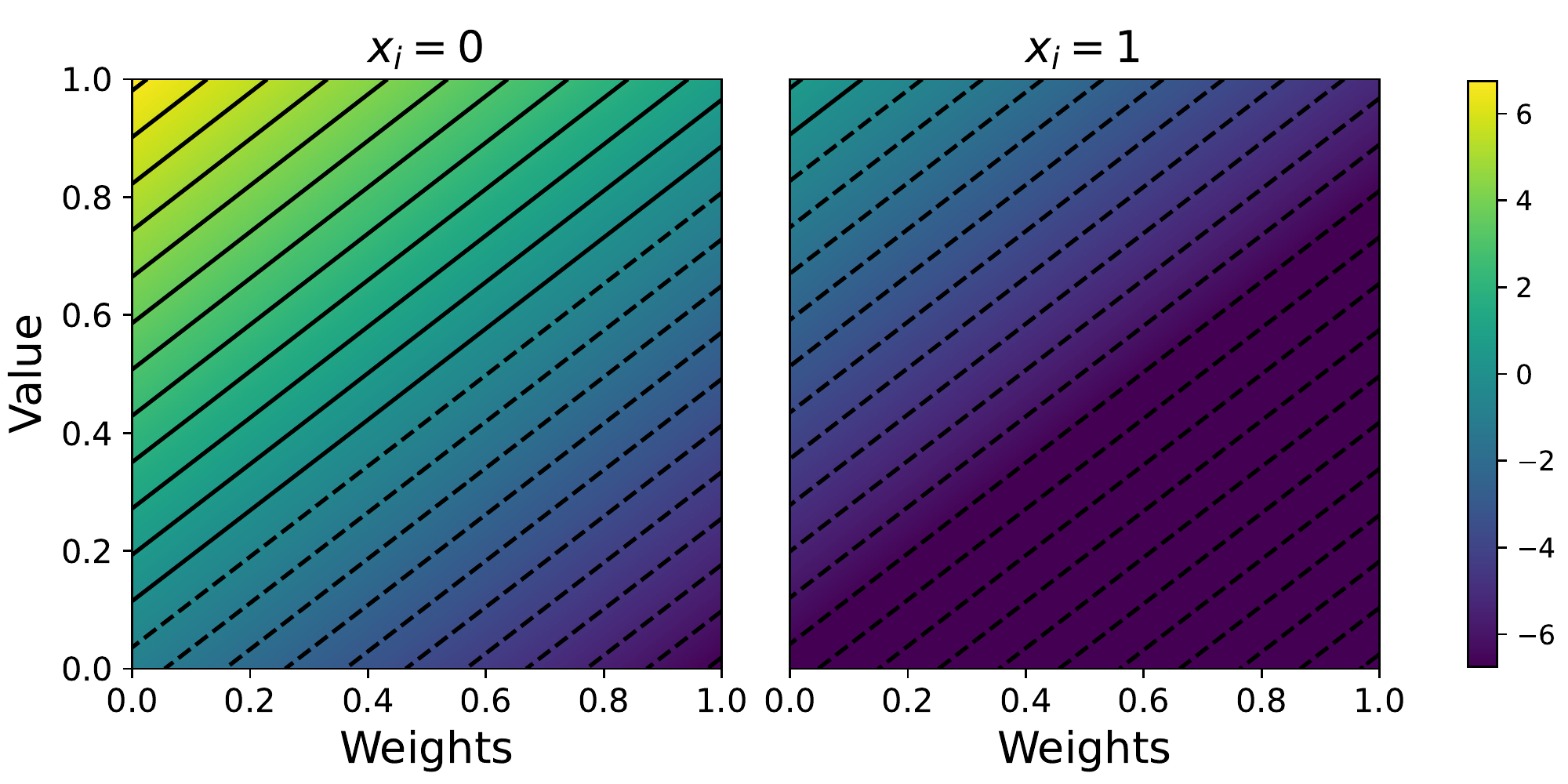}
    \caption{Knapsack Policy with logits for $x_i=0$ and $x_i=1$ shown in each pane. Light valuable objects are favoured to insert. Once inserted the policy downweights an object's probably of flipping state again. Interestingly, the ejection probability of heavy, valueless objects is low, perhaps because this only makes sense close to overflowing, although the policy does not receive free capacity as a feature.}
    \label{fig:knapsack-ablation}
\end{figure}

\begin{table*}
    \small
    \caption{Average cost of solutions for the Knapsack Problem across five random seeds and, in parentheses, optimality gap to best solution found among solvers. Bigger is better. *Values as reported by \cite{bello16} for reference.}
    \label{tab:knapsack-results}
    \centering
    \resizebox{\textwidth}{!}{
    \begin{tabular}{c | c c c | c  c  c | c c}
    \toprule
        & Random Search & Bello RL & Bello AS & SA & Ours (PPO) & Ours (ES) & Greedy & OR-Tools \\
    \midrule
        \textsc{Knap50}     & $17.91^*$  & $19.86^*$ & $20.07^*$ & $18.43\,(8.40\%)$ & $19.69\,(2.23\%)$ & $19.95\,(0.84\%)$ &  $19.94\,(0.89\%)$ & $\bm{20.12}\,(0.00\%)$ \\
        \textsc{Knap100}    & $33.23^*$ &  $40.27^*$ & $40.50^*$ & $36.81\,(8.91\%)$ & $39.54\,(2.15\%)$ & $39.90\,(1.26\%)$ & $40.17\,(0.59\%)$ & $\bm{40.41}\,(0.00\%)$ \\
        \textsc{Knap200}    & $35.95^*$ &  $57.10^*$ & $57.45^*$ & $50.89\,(11.73\%)$ & $55.03\,(4.54\%)$ & $55.58\,(3.59\%)$ & $57.30\,(0.61\%)$ & $\bm{57.65}\,(0.00\%)$ \\
        \textsc{Knap500}    &  & - & - & $126.92\,(11.95\%)$ & $138.14\,(4.16\%)$ & $141.01\,(2.17\%)$ & $143.77\,(0.25\%)$ & $\bm{144.14}\,(0.00\%)$ \\
        \textsc{Knap1K}    & - &  - & - & $254.45\,(11.96\%)$ & $277.41\,(4.01\%)$ & $282.46\,(2.26\%)$ & $288.64\,(0.13\%)$ & $\bm{289.01}\,(0.00\%)$ \\
        \textsc{Knap2K}    & - &  - & - & $507.72\,(12.03\%)$ & $554.32(3.97\%)$ & $563.75(2.34\%)$ & $576.89\,(0.06\%)$ & $\bm{577.28}\,(0.00\%)$ \\
    \bottomrule
    \end{tabular}
    }
\end{table*}

Solutions are represented as a binary vector $\rvx$, with $x_i=0$ for `out of the bin' and $x_i=1$ for `in the bin'. Our proposal distribution flips individual bits, one at a time, with the constraint that we cannot flip $0 \mapsto 1$ if the bin capacity will be exceeded. The neighbourhood of $\rvx_k$ is thus all feasible solutions at a Hamming distance of 1 from $\rvx_k$. We use the proposal distribution described in Section~\ref{sec:arch} and illustrated in Figure~\ref{fig:policy-net}, consisting of a pointwise embedding of each item---its weight, value, occupancy bit, the knapsack's overall capacity, and global temperature---into a logit-space, followed by a softmax. Mathematically the policy and state--action to proposal mapping are
\begin{align}
    \nonumber \pi_\theta(i | \rvs) &= \text{softmax}\left ( \rvz \right)_i, \enspace z_i = f_\theta([x_i, w_i, v_i, W, T]) \\
    \rvx' &= \rvx + \text{onehot}(i)\mod 2.
\end{align}
where $f_\theta$ is a small two-layer neural network $5 {\to} 16 {\to} 1$ with ReLU activations, comprising only 112 parameters. Actions are sampled from the categorical distribution induced by the softmax and cast to one-hot vectors $\text{onehot}(i)$.

Neural networks have been used to solve the Knapsack Problem in \cite{vinyals2017pointer}, \cite{nomer2020neural}, and \cite{bello16}. We follow the setup of \citet{bello16}, honing in on 3 self-generated datasets: \textsc{Knap50}, \textsc{Knap100} and \textsc{Knap200}. \textsc{Knap}$N$ consists of $N$ items with weights and values generated uniformly at random in $(0,1]$ and capacities $C_{50} \!=\! 12.5, C_{100} \!=\! 25$, and $C_{200} \!=\! 25$. We use OR-Tools \citep{ortools} to compute ground truth solutions. Results in Table~\ref{tab:knapsack-results} show that Neural SA improves over vanilla SA by up to 10\% optimality gap, and heuristic methods (Random Search) by much more. Neural SA falls slightly behind two methods by \citet{bello16}, which use (1) a large attention-based pointer network with several orders of magnitude more parameters in Bello RL, and (2) this coupled with 5000 iterations of their Active Search method. It also falls behind a greedy heuristic for packing a knapsack based on the value-to-weight ratio. In Figure~\ref{fig:knapsack-ablation} we analyse the policy network and a typical rollout. It has learnt a mostly greedy policy to fill its knapsack with light, valuable objects, only ejecting them when full. This is in line with the value-to-weight greedy heuristic. Despite not coming top among methods, we note Neural SA is typically within 1-3\% of the minimum energy, although its architecture was not designed for this problem in particular.

\subsection{Bin Packing Problem} \label{sec:bin-packing}
\begin{figure}[!htb]
    \small
    \centering
    \includegraphics[width=.7\columnwidth]{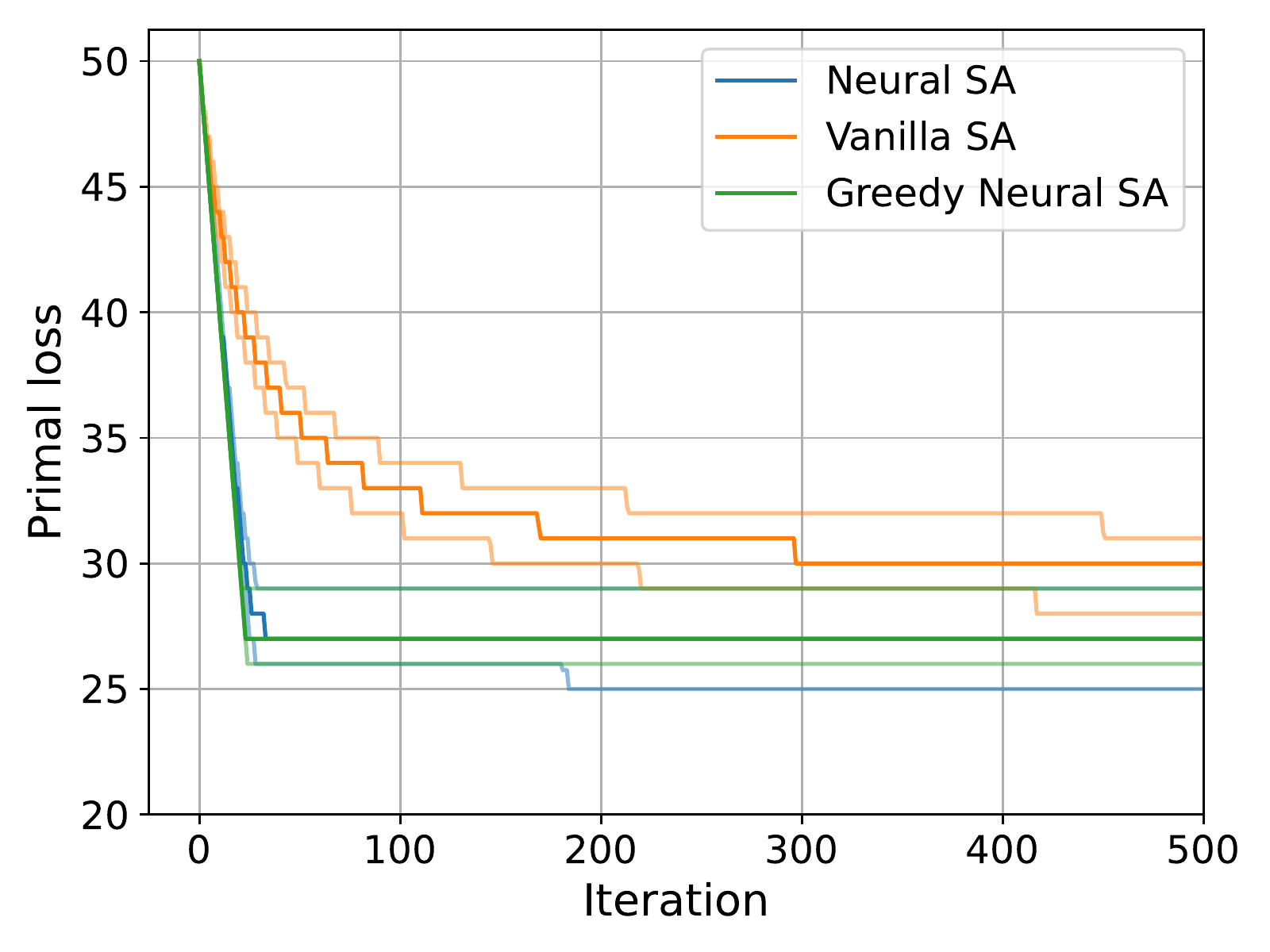}
    \caption{\textsc{Bin50} primal objective for vanilla, Neural, and Greedy Neural SA with 25\textsuperscript{th}, 50\textsuperscript{th}, and 75\textsuperscript{th} percentiles. 
    }
    \label{fig:bin-packing-primal}
\end{figure}

\begin{figure}[!htb]
    \small
    \includegraphics[width=\columnwidth]{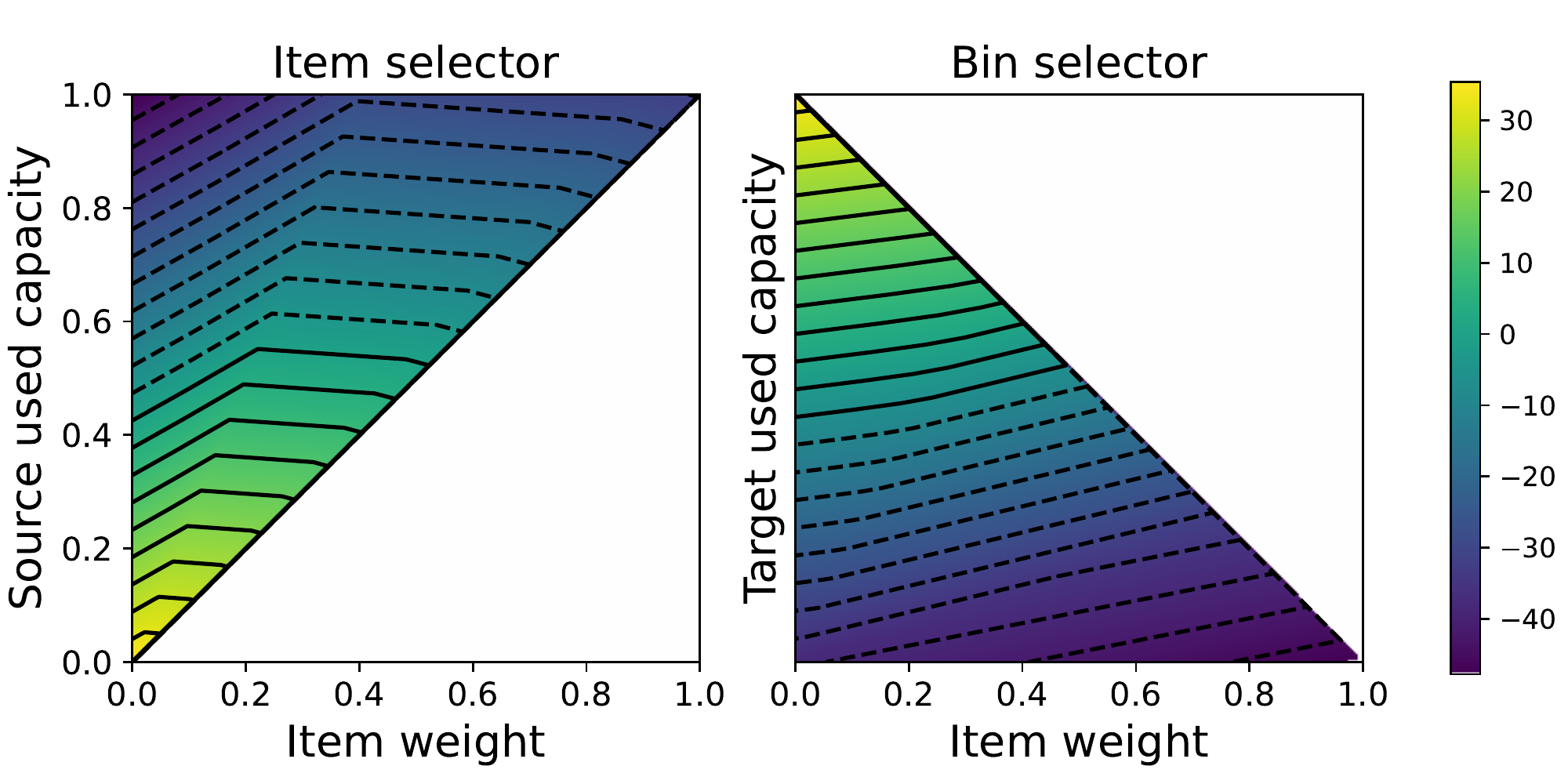}
    \caption{Bin Packing policy (logits), consisting of two networks, an item selector and a bin selector. The item selector uses item weight and bin used capacities to select an item to move. The bin selector then places this item in a bin, based on target bin fullness and selected item weight. The learnt policy is very sensible. The item selector looks for a light item in an under-full bin. The bin selector then place this in an almost-full bin. We mask bins with insufficient free capacity, hence the triangular logit-spaces.}
    \label{fig:binpacking-logits}
\end{figure}

The Bin Packing problem is similar to the Knapsack problem in nature. Here, one wants to pack \emph{all} of $N$ items into the smallest number of bins possible, where each item $i \in \{1, \cdots, N\}$ has weight $w_i$, and we assume, without loss of generality, $N$ bins of equal capacity $W \geq \max_i(w_i)$; there would be no valid solution otherwise. This problem is NP-hard and has a search space of size equal to the $N^{th}$ Bell number.
If $x_{ij}$ denotes item $i$ occupying bin $j$, then the problem can be written as minimising an energy:
\begin{align}
    \text{minimise} \enspace & E(\rvx; \bm{\psi}) = \sum_{j=0}^{N-1} y_j, \\
    \nonumber \text{subject to }  & \underbrace{\sum_{i=0}^{N-1} w_i x_{ij} \leq W }_{\text{bin capacity constraint}}, \quad \underbrace{\sum_{j=0}^{N-1} x_{ij} = 1}_{\text{1 bin per item}}, \\ 
    \nonumber & \underbrace{y_j = \min \left (1, \sum_{i=0}^{N-1} x_{ij} \right)}_{\text{bin occupancy indicator}}, \quad x_{ij} \in \{0,1\}
\end{align}
where the constraints apply for all $i$ and $j$. We define the policy in two steps: we first pick an item $i$, and then select a bin $j$ to place it into.
We can then write the policy as $\pi_{\theta, \phi}(\rva{=}(i,j)| \rvs) {=} \pi_\phi(i | \rvs)\pi_\theta(j | \rvs, i)$, which we define as
\begin{align}
    \nonumber \pi_\theta(i | \rvs) {=} \text{softmax}\left ( \rvz^{\text{item}} \right)_i &, \enspace z^{\text{item}}_i {=} f_\theta([w_i, c_{b(i)}, T]), \\
    \pi_\phi(j | \rvs, i) {=} \text{softmax} \left ( \rvz^{\text{bin}} \right)_j &, \enspace z^{\text{bin}}_j {=} f_\phi([w_i, c_j, T]),
\end{align}
where $b(i)$ is the bin item $i$ is in before the action (in terms of $x_{ij}$, we have $x_{ib(i)} = 1$), $c_j$ is the free capacity of bin $j$ ($c_j = W - \sum_{i=1}^N w_i x_{ij}$), and both $f_\theta$ and $f_\phi$ are lightweight architectures $3 {\to} 16 {\to} 1$ with a ReLU nonlinearity between the two layers. We sample from the policy ancestrally, sampling first an item from $\pi_\theta(i | \rvs)$, followed by a bin from $\pi_\phi(j | \rvs, i)$. Results in Table~\ref{tab:binpacking-results} show that our lightweight model is able to find a solution to about 1\% higher energy than the minimum found by FFD \cite{johnson1973near}, a very strong heuristic for this problem \citep{rieck2021basic}. We even see that we very often beat the SCIP \citep{GamrathEtal2020OO,GamrathEtal2020ZR} optimizer in OR-Tools, which timed out on most problems. Figure~\ref{fig:bin-packing-primal} compared convergence speed of Neural SA with vanilla SA and a third option, Greedy Neural SA, which uses argmax samples from the policy. The learnt policy, visualised in Figure~\ref{fig:binpacking-logits} has much faster convergence than the vanilla version. Again, we see that our method, although simple, is competitive with hand-designed alternatives, whereas vanilla SA is not.



\begin{table*}
    \small
    \caption{Average cost of solutions for the Bin Packing Problem across five random seeds and, in parentheses, optimality gap to best solution found among solvers. Lower is better. We set a time out for Or-Tools of 1 minute per problem for \textsc{Bin50-1000} and of 2 minutes for \textsc{Bin2000}; * indicates only the trivial solution was found in this time.}
    \label{tab:binpacking-results}
    \centering
    \begin{tabular}{l c c c | c c }
    \toprule
        & SA & Ours (PPO) & Ours (ES) & OR-Tools (SCIP) & FFD \\
    \midrule
        \textsc{Bin50}  & $30.38\,(13.74\%)$ & $27.32\,(2.28\%)$ & $27.24\,(1.98\%)$ & $\bm{26.71}\,(0.00\%)$ & $27.10\,(1.46\%)$ \\
        \textsc{Bin100} & $60.66\,(14.65\%)$ & $53.53\,(1.17\%)$ & $53.38\,(0.88\%)$ & $53.91\,(1.89\%)$ & $\bm{52.91}\,(0.00\%)$ \\
        \textsc{Bin200} & $121.27\,(16.32\%)$ & $105.63\,(1.32\%)$ & $105.43\,(1.13\%)$ & $109.19\,(4.74\%)$ & $\bm{104.25}\,(0.00\%)$ \\
        \textsc{Bin500} & $302.84\,(17.82\%)$ & $259.08\,(0.80\%)$ & $259.09\,(0.80\%)$ & $267.63\,(4.13\%)$ & $\bm{257.02}\,(0.00\%)$ \\
        \textsc{Bin1000} & $605.23\,(18.79\%)$ & $512.66\,(0.63\%)$ & $512.66\,(0.63\%)$ & $1000^*$ & $\bm{509.46}\,(0.00\%)$ \\
        \textsc{Bin2000} & $1209.72\,(18.84\%)$ & $\bm{1017.88}\,(0.00\%)$ & $\bm{1017.88}\,(0.00\%)$ & $2000^*$ & $1028.67\,(1.06\%)$ \\
    \bottomrule
    \end{tabular}
\end{table*}

\subsection{Travelling Salesperson Problem} \label{sec:tsp}

Imagine you will make a round road-trip through $N$ cities and want to plan the shortest route visiting each city once; this is the Travelling Salesperson Problem (TSP) \citep{applegate2006traveling}. The TSP has been a long time favourite of computer scientists due to its easy description and NP-hardness (the base search space has size equal to the factorial of the number of cities). Here we use it as an example of a difficult CO problem. We compare with Concorde \citep{applegate2006traveling} and LKH-3 \citep{helsgaun2000effective}, two custom solvers for TSP. Given cities $i \in \{0,1,...,N-1\}$ with spatial coordinates $\rvc_{i} \in [0, 1]^2$, we wish to find a linear ordering of the cities, called a \emph{tour}, denoted by the permutation vector $\rvx = (x_0, x_1, ..., x_{N-1})$ for $x_i \in \{0,1,...,N-1\}$ such that
\begin{align}
    & \nonumber \text{minimise } E(\rvx; \bm{\psi}) = \sum_{i=0}^{N-1} \Vert \rvc_{x_{i+1}} - \rvc_{x_i} \Vert_2 \\
    & \text{subject to }  x_i \neq x_j \text{ for all } i \neq j \\
    \nonumber & \qquad \, \text{ and } x_i \in \{0,1,...,N-1\},
\end{align}
where we have defined $x_N = x_0$ for convenience of notation. Our action space consists of so-called \emph{2-opt} moves \citep{croes1958method}, which reverse contiguous segments of a tour. An example of a 2-opt move is shown in Figure~\ref{fig:neural_sa_pipeline}. We have a two-stage architecture, like in Bin Packing, which selects the start and end cities of the segment to reverse. Denoting $i$ as the start and $j$ as the end cities, we have $\pi_{\theta, \phi}(\rva {=} (i,j)|\rvs) = \pi_\phi(i | \rvs) \pi_\theta(j | \rvs, i)$, parametrised as
\begin{align}
    \nonumber \pi_\theta(i | \rvs) &= \text{softmax}\left ( \rvz \right)_i,
    \enspace z_i = f_\theta([\rvc_{\rvx_{[i-1:i+1]}}, T]), \\
    \nonumber \pi_\phi(j | \rvs, i) &= \text{softmax}\left ( \rvz \right)_j, \\
    z_j &= f_\phi([\rvc_{\rvx_{[i{-}1:i{+}1]}}, \rvc_{\rvx_{[j{-}1:j{+}1]}}, T]),
\end{align}
where $\rvx_{[i-1:i+1]}$ are the indices of city $i$ and its tour neighbours $i-1$ and $i+1$. Again, we use simple MLPs: $f_\theta$ has architecture $7 {\to} 16 {\to} 1$ and $f_\phi$, $13 {\to} 16 {\to} 1$. We test on publicly available TSP20/50/100 \citep{kool2018attention} with 10K problems each and generate TSP200/500 with 1K tours each. Results, in Table~\ref{tab:TSP-full}, show Neural SA improves on vanilla SA. Albeit not outperforming Fu et al.~\yrcite{fu2021generalize}, Neural SA is neck-to-neck with other neural improvement heuristics methods, GAT-T\{1000\} \citep{wu2021learning} and Costa\{500\} \citep{costa2020}. Since Neural SA is not custom designed for TSP as the competing methods, we view this as surprisingly good. A more complete comparison, including other neural approaches, is given in the appendix, Table~\ref{tab:TSP-full}.

\section{Discussion}
\begin{table*}[!htb]
    \small
    \caption{Comparison of Neural SA against competing methods with similar running times on TSP. Extended version in Table~\ref{tab:TSP-full}. Lower is better. *Values as reported in respective works \citep{wu2019learning,costa2020,fu2021generalize}.}
    \label{tab:tsp}
    \centering
    \resizebox{\textwidth}{!}{
    \begin{tabular}{l | r r r | r r r | r r r | rrr | rrr }
    \toprule
        &  \multicolumn{3}{| c |}{TSP20} & \multicolumn{ 3 }{| c |}{ TSP50 } & \multicolumn{ 3 }{| c }{TSP100} & \multicolumn{ 3 }{| c |}{ TSP200 } & \multicolumn{ 3 }{| c }{ TSP500 }\\
        & Cost & Gap & Time & Cost & Gap & Time & Cost & Gap & Time & Cost & Gap & Time & Cost & Gap & Time \\
    \midrule
        \textsc{Concorde}  & 3.836 & 0.00\% & 48s & 5.696 & 0.00\% & 2m & 7.764 & 0.00\% & 7m & 10.70 & 0.00\% & 38m & 16.54 & 0.00\% & 7h58m \\
        \textsc{LKH-3}     & 3.836 & 0.00\% & 1m & 5.696 & 0.00\% & 14m & 7.764 & 0.00\% & 1h & 10.70 & 0.00\% & 21m & 16.54 & 0.00\% & 1h15m \\
    \midrule
        \textsc{SA}  & 3.881 & 1.17\% & 5s & 5.943 & 4.34\% & 37s & 8.343 & 7.45\% & 3m & 11.98 & 11.87\% & 9m & 20.22 & 22.25\% & 56m \\
        \textsc{Ours (PPO)}    & 3.838 & 0.05\% & 9s & 5.734 & 0.67\% & 1m & 7.874 & 1.42\% & 9m & 11.00 & 2.80\% & 16m & 17.64 & 6.65\% & 2h16m  \\
        \textsc{Ours (ES)}    & 3.840 & 0.10\% & 9s & 5.828 & 2.32\% & 1m & 8.191 & 5.50\% & 9m & 11.74 & 9.72\% & 16m & 20.27 & 22.55\% & 2h16m  \\
    \midrule
        \textsc{OR-Tools}*     & 3.86 & 0.85\% & 1m & 5.85 & 2.87\% & 5m & 8.06 & 3.86\% & 23m & - & - & - & - & - & - \\
        \textsc{GAT-T}\{1000\}* & 3.84 & 0.03\% & 12m & 5.75 & 0.83\% & 16m & 8.01 & 3.24\% & 25m & - & - & - & - & - & -  \\
        Costa \{500\}* & 3.84 & 0.01\% & 5m & 5.72 & 0.36\% & 7m & 7.91 & 1.84\% & 10m & - & - & -  & - & - & -  \\
        Fu et al.* & 3.84 & 0.00\% & 1m & 5.70 & 0.01\% & 8m & 7.76 & 0.04\% & 15m & - & - & - & - & - & -  \\
    \bottomrule
    \end{tabular}
    }
\end{table*}
Neural SA is a general, plug-and-play method, requiring only the definition of neighbourhoods for the proposal distribution and training problem instances (no solutions needed). It also obviates time-consuming architecture design, since a simple MLP is enough for a range of CO problems. In this section, we discuss some of the main features of Neural SA.

\textbf{Computational Efficiency} 
Neural SA requires little computational resources given its compact architecture, with 384 parameters on TSP, 160 for Bin Packing, and 112 for Knapsack. Further, the cost of each step scales linearly in the problem size, since the architectures are embarrassingly parallel.
In terms of running times, Neural SA is on par with and often faster than other TSP solvers (see Tables~\ref{tab:tsp}, \ref{tab:TSP-full}). For the Knapsack and Bin Packing problems, we compare running times against OR-Tools, as shown in the appendix, Table~\ref{tab:times}. Neural SA lags behind OR-Tools in the Knapsack, for which an efficient branch and bound solver is known. However, for the Bin Packing problem, Neural SA is much faster than the available Mixed-Integer Programming solver, which only found trivial solutions for $N{\geq}1000$. Finally, Neural SA is also fast to train; only a few minutes with PPO and up to a few hours with ES. This is can be attributed to its low number of parameters but also to its generalisation ability; in all experiments, we could get away with training \emph{only on the smallest instances with very short rollouts}.

\textbf{PPO vs ES} 
Neural SA can be trained with any policy optimisation method making it highly extendable. We found no winner between PPO and ES, apart from on the TSP, where PPO excelled and generalized better to larger instances. We also observed PPO to converge $\sim10\times$ faster than ES, but ES policies were more robust, still performing well when we switched to greedy sampling, for example.
Interestingly, the acceptance rate over trajectories was problem dependent and always higher in Neural SA (both PPO and ES) than in vanilla SA, contradicting conventional wisdom that it should be held at 0.44 throughout a rollout \citep{lam1988performance}.

\textbf{Generalisation}
Our experiments show Neural SA generalises to different problem sizes and rollout lengths; a remarkable feat for such a simple pipeline, since transfer learning is notoriously difficult in RL and CO. Many ML4CO methods do handle problems of different sizes but underperform when tested on larger instances than the ones seen in training \cite{kool2018attention,joshi2019learning} (see appendix, Table~\ref{tab:generalisation-results}).
Fu et al.~\yrcite{fu2021generalize} achieve better generalisation results for the TSP but had to resort to a suite of techniques to allow a small supervised model to be applied to larger problems. These are not easy to implement, TSP-specific, and consist only the first step in a complex pipeline that still relies on a tailored Monte-Carlo tree search algorithm.

\textbf{Solution Quality} 
In all problems we considered, Neural SA, with little to no fine-tuning of its hyperparameters, outperformed vanilla SA and could get within a few percentage points or less of global minima. Conversely, state-of-the-art SA variants are designed by searching a large space of different hyperparameters \cite{franzin2019revisiting}, a costly process that Neural SA helps us mitigate.
Neural SA did not achieve state-of-the-art results, but that was not to be expected nor our main goal. Instead, we envision Neural SA as a general purpose solver, allowing researchers and practitioners to get a strong baseline quickly without the need to fine-tune classic CO algorithms or design and train complex neural networks.
Given the good performance, small computational resources, and fast training across a diverse set of CO problems, we believe Neural SA is a promising solver that can strike the right balance among solution quality, computing costs and development time.

\section{Conclusion}
We presented \emph{Neural SA}, neurally augmented simulated annealing, where the SA chain is a trajectory from an MDP. In this light, the proposal distribution could be interpreted as a policy, which could be optmised. This has numerous benefits: 1) accelerated convergence of the chain, 2) ability to condition the proposal distribution on side-information 3) no need of ground truth data to learn the proposal distribution, 4) lightweight architectures that can be run on CPU unlike many contemporary ML4CO methods, 5) scalability to large problems due to its lightweight computational overhead, 6) generalisation across different problem sizes. 

These contributions show augmenting classic, time-tested (meta-)heuristics with learnable components is a promising direction for future research in ML4CO. In contrast to expensive end-to-end methods in previous work, this could be a more promising path towards machine learning models capable of solving a wide range of CO problems. As we show in this paper, this approach can yield solid results for different problems while preserving theoretical guarantees of existing CO algorithms and requiring only simple neural architectures that can be easily trained on small problems.

The ease of use and flexibility of Neural SA do come with drawbacks. In all experiments we were not able to achieve the minimum energy, even though we could usually get within a percentage point. The model also has no built-in termination condition, neither can it provide a certificate on the quality of solutions found. There is still also the question of how to tune the temperature schedule, which we did not attempt in this work. These shortcomings are all points to be addressed in upcoming research. We are also interested in extending the framework to multiple trajectories, such as in parallel tempering \cite{swendsenwang1986} or genetic algorithms \cite{holland1993}. For these, we would maintain a population of chains, which could exchange information.

\bibliography{icml2022_conference}
\bibliographystyle{icml2022}

\clearpage
\appendix
\section{Additional Experimental Information and Results}
\subsection{General Information}
\paragraph{Implementation} Our code was implemented in Pytorch 1.9 \citep{paszke2017automatic} and run in a standard machine with a single GPU RTX2080. The code will be made publicly available upon publication.

\paragraph{Architectures} In all experiments, the proposal distribution is parametrised by a two-layer neural network, with ReLU activation and 16 neurons in the hidden layer: $\text{input\_size} \to 16 \to 1$, where the size of the input is problem specific. When using PPO, we also need a critic network to estimate the state-value function so that we can compute advantages using Generalised Advantage Estimator (GAE) \citep{schulman2016high}. The critic network does not share any parameters with the proposal distribution (actor) but has the exact same architecture. The only difference is that the actor outputs logits of the proposal distribution, whereas the critic outputs action values from which we compute the necessary state values.

\paragraph{Training}
We train Neural SA using both Proximal Policy Optimisation (PPO) \citep{schulman2017proximal} and Evolution Strategies (ES) \citep{salimans2017evolution}. Across all experiments, most of the hyper-parameters of both of these methods are kept constant, as detailed below.
\begin{itemize}[itemsep=0pt,topsep=0pt]
    \item \textbf{PPO}: We optimise both actor and critic networks using Adam \citep{kingma2015adam} with learning rate of $2\mathrm{e}{-4}$, weight decay of $1\mathrm{e}{-2}$ and $\beta = (0.9, 0.999)$. For PPO, we set the discount factor and clipping threshold to $\gamma=0.9$ and $\epsilon=0.25$, respectively, and compute advantages using GAE \citep{schulman2016high} with trace decay $\lambda=0.9$.
    \item \textbf{ES}: We use a population of 16 perturbations sampled from a Gaussian of standard deviation 0.05. Updates are fed into an SGD optimizer with learning rate 1e-3 and momentum 0.9.
\end{itemize}

\paragraph{Testing} The randomly generated datasets used for testing can be recreated by setting the seed of Pytorch's random number generator to $0$. Similarly, we evaluate each configuration (problem size, number of steps) 5 times and report the average as well as the standard deviation across the different runs. For reproducibility, we also seed each of these runs (seeds $1$, $2$, $3$, $4$ and $5$).

\paragraph{Running Times}
We compare the running times of Neural SA and other combinatorial optimisation methods. Table~\ref{tab:times} shows the running times of Neural SA against those of OR-Tools at the Knapsack and Bin Packing problems, while Table~\ref{tab:TSP-full} show running times on the Travelling Salesperson Problem for Neural SA and a number of competing solvers.

\begin{table}[!htb]
    \small
    \caption{Comparison of running times (at test time) for Neural SA (PPO/ES) against OR-tools for the Knapsack and Bin Packing Problems. We report the average time to evaluate one instance with each method for different problem sizes.}
    \label{tab:times}
    \centering
    \begin{tabular}{l | r r | r r }
    \toprule
        & \multicolumn{2}{| c |}{Knapsack} & \multicolumn{2}{| c }{Bin Packing}\\
        & Ours & OR-Tools & Ours & OR-Tools \\
    \midrule
        \textsc{50N}  & $<1s$ & $<1s$ & $<1s$ & $54s$ \\
        \textsc{100N} & $1s$ & $<1s$ & $1s$ & $56s$ \\
        \textsc{200N} & $2s$ & $<1s$ & $3s$ & $\geq1m$ \\
        \textsc{500N} & $6s$ & $1s$ & $10s$ & $\geq1m$ \\
        \textsc{1000N} & $18s$ & $2s$ & $29s$ & $\geq1m$ \\
        \textsc{2000N} & $1m5s$ & $8s$ & $1m43s$ & $\geq2m$ \\
    \bottomrule
    \end{tabular}
\end{table}

\subsection{Knapsack Problem}
\paragraph{Data} We consider different problem sizes, with \textsc{Knap}$N$ consisting of $N$ items, each with a weight $w_i$ and value $v_i$ sampled from a uniform distribution, $w_i, v_i \sim \gU_{(0:1)}$. Each problem has also an associated capacity, that is, the maximum weight the knapsack can comport. Here we follow \citep{bello16} and set $C_{50}{=}12.5, C_{100}{=}25$ and $C_{200}{=}25$. However, for larger problems we set $C_N=N/8$.

\paragraph{Initial Solution} We start with a feasible initial solution corresponding to an empty knapsack, that is, $\rvx=\mathbf{0}$. That is the trivial (and worst) feasible solution, so our models do not require any form of initialisation pre-processing or heuristic.

\paragraph{Training} We train \emph{only on} \textsc{Knap50} with short rollouts of length $K=100$ steps. The model is trained for 1000 epochs each of which is run on 256 random problems generated on the fly as described in the previous section. We set initial and final temperatures to $T_0=1$ and $T_K=0.1$, and compute the temperature decay as $\alpha = \left(T_K / T_0 \right)^{\frac{1}{K}}$.

\paragraph{Testing} We evaluate Neural SA on test sets of 1000 randomly generated Knapsack problems, while varying the length of the rollout. For each problem size $N$, we consider rollouts of length $K=N$, $K=2N$, $K=5N$ and $K=10N$. The initial and final temperatures are kept fixed to $T_0=1$ and $T_K=0.1$, respectively, and the temperature decay varies as function of $K$, $\alpha = \left(T_K / T_0 \right)^{\frac{1}{K}}$. 

We compare our methods against one of the dedicated solvers for knapsack in OR-Tools \citep{ortools} (Knapsack Multidimension Branch and Bound Solver). We also compare sampled and greedy variants of Neural SA. The former samples actions from the proposal distribution while the latter always selects the most likely action.

\begin{table*}[!htb]
    \small
    \caption{ES results on the Knapsack benchmark. Bigger is better. Comparison among rollouts of different lengths: 1, 2, 5 or 10 times the dimension of the problem.}
    \label{tab:knapsack-full-es}
    \centering
    \begin{tabular}{l | r | rrrr || r}
    \toprule
        & \multicolumn{1}{| c |}{Greedy} & \multicolumn{4}{| c ||}{Sampled} & OR-Tools  \\
        & $\times 1$ & $\times 1$ & $\times 2$ & $\times 5$ & $\times 10$ & \\
    \midrule
        \textsc{Knap50}     & $16.59\pm.00$ & $19.45\pm.01$ & $19.70\pm.00$ & $19.86\pm.00$ & $19.95\pm.00$ & $\bm{20.12}$ \\
        \textsc{Knap100}    & $31.15\pm.00$ & $39.07\pm.01$ & $39.49\pm.01$ & $39.76\pm.01$ & $39.90\pm.01$ & $\bm{40.41}$ \\
        \textsc{Knap200}    & $55.96\pm.00$ & $53.72\pm.02$ & $55.21\pm.02$ & $56.22\pm.02$ & $56.58\pm.01$ & $\bm{57.65}$ \\
        \textsc{Knap500}    & $135.92\pm.00$ & $134.20\pm.05$ & $137.89\pm.03$ & $140.20\pm.02$ & $141.01\pm.03$ & $\bm{144.14}$ \\
        \textsc{Knap1K}   & $259.20\pm.00$ & $269.21\pm.04$ & $276.48\pm.05$ & $280.94\pm.02$ & $282.46\pm.03$ & $\bm{289.01}$ \\
        \textsc{Knap2K}   & $489.02\pm.00$ & $537.53\pm.08$ & $551.92\pm.07$ & $560.75\pm.07$ & $563.75\pm.02$ & $\bm{577.28}$ \\
    \bottomrule
    \end{tabular}
\end{table*}

\begin{table*}[!htb]
    \small
    \caption{PPO results on the Knapsack benchmark. Bigger is better. Comparison among rollouts of different lengths: 1, 2, 5 or 10 times the dimension of the problem.}
    \label{tab:knapsack-full-ppo}
    \centering
    \begin{tabular}{l | r | rrrr || r}
    \toprule
        & \multicolumn{1}{| c |}{Greedy} & \multicolumn{4}{| c ||}{Sampled} & OR-Tools \\
        & $\times 1$ & $\times 1$ & $\times 2$ & $\times 5$ & $\times 10$ & \\
    \midrule
        \textsc{Knap50}     & $19.52\pm.00$ & $19.37\pm.01$ & $19.42\pm.01$ & $19.55\pm.01$ & $19.69\pm.01$ & $\bm{20.12}$ \\
        \textsc{Knap100}    & $38.97\pm.00$ & $38.64\pm.01$ & $38.81\pm.01$ & $39.20\pm.01$ & $39.54\pm.01$ & $\bm{40.41}$ \\
        \textsc{Knap200}    & $48.58\pm.00$ & $48.99\pm.06$ & $51.00\pm.04$ & $53.57\pm.03$ & $55.03\pm.01$ & $\bm{57.65}$ \\
        \textsc{Knap500}    & $119.38\pm.00$ & $122.40\pm.03$ & $128.46\pm.05$ & $134.95\pm.03$ &  $138.14\pm.04$ & $\bm{144.14}$ \\
        \textsc{Knap1K}     & $238.18\pm.00$ & $246.54\pm.09$ & $259.15\pm.08$ & $271.68\pm.05$ & $277.41\pm.03$ & $\bm{289.01}$ \\
        \textsc{Knap2K}     & $472.67\pm.00$ & $493.47\pm.07$ & $519.27\pm.08$ & $543.72\pm.11$ & $554.32\pm.04$ & $\bm{577.28}$ \\
    \bottomrule
    \end{tabular}
\end{table*}

\subsection{Bin Packing Problem}

\paragraph{Data} We consider problems of different sizes, with \textsc{Bin}$N$ consisting of $N$ items, each with a weight (size) sampled from a uniform distribution, $w_i \sim \gU_{(0:1)}$. Without loss of generality, we also assume $N$ bins, all with unitary capacity. Each dataset \textsc{Bin}$N$ in Tables~\ref{tab:binpacking-full-es} and \ref{tab:binpacking-full-ppo} contains 1000 such random Bin Packing problems used to evaluate the methods at test time.

\paragraph{Initial Solution} We start from the solution where each item is assigned to a different bin, e.g. $x_{ij} = i$.

\paragraph{Training} We train \emph{only on} \textsc{Bin50} with short rollouts of length $K=100$ steps. The model is trained for 1000 epochs each of which is ran on 256 random problems generated on the fly as described in the previous section. We keep the same temperature decay with $\alpha = \left(T_K / T_0 \right)^{\frac{1}{K}}$, but use different initial and final temperatures for PPO and ES. For PPO, we set $T_0=1$ and $T_K=0.1$, whereas for ES we set $T_0=0.1$ and $T_K=1e-4$.

\paragraph{Testing} We evaluate Neural SA on test sets of 1000 randomly generated Bin Packing problems, while varying the length of the rollout. For each problem size $N$, we consider rollouts of length $K=N$, $K=2N$, $K=5N$ and $K=10N$. The initial and final temperatures are kept the same as in training, and the temperature decay parameter varies as function of $K$, $\alpha = \left(T_K / T_0 \right)^{\frac{1}{K}}$. 

We compare Neural SA against First-Fit-Decreasing (FFD) \citep{johnson1973near}, a powerful heuristic for the Bin Packing problem, and against OR-Tools \citep{ortools} MIP solver powered by SCIP \citep{GamrathEtal2020OO}. The OR-Tools solver can be quite slow on Bin Packing so we set a time out of 1 minute per problem for BIN50-1000 and of 2 minutes for BIN2000 to match Neural SA running times (see Table~\ref{tab:times}).

We also compare sampled and greedy variants of Neural SA. The former naturally samples actions from the proposal distribution while the latter always selects the most likely action.

\begin{table*}[ht]
    \small
    \caption{ES results on the Bin Packing benchmark. Lower is better.}
    \label{tab:binpacking-full-es}
    \centering
    \begin{tabular}{l | r | rrrr || rr}
    \toprule
        & \multicolumn{1}{| c |}{Greedy} & \multicolumn{4}{| c ||}{Sampled} & OR-Tools & FFD \\
        & $\times 1$ & $\times 1$ & $\times 2$ & $\times 5$ & $\times 10$ & \\
    \midrule
        \textsc{Bin50}  & 27.62$\pm.00$ & 27.43$\pm.01$ & 27.36$\pm.01$ & 27.29$\pm.00$ & 27.24$\pm.01$ & $\bm{26.71}$ & 27.10 \\
        \textsc{Bin100} & 53.80$\pm.00$ & 53.63$\pm.00$ & 53.54$\pm.01$ & 53.44$\pm.01$ & 53.38$\pm.01$ & 53.91 & $\bm{52.91}$ \\
        \textsc{Bin200} & 105.63$\pm.00$ & 105.78$\pm.02$ & 105.64$\pm.01$ & 105.51$\pm.01$ & 105.43$\pm.01$ & 109.19 & $\bm{104.25}$ \\
        \textsc{Bin500} & 259.09$\pm.00$ & 260.86$\pm.03$ & 260.65$\pm.01$ & 260.42$\pm.02$ & 260.27$\pm.02$ & $267.63$ & $\bm{257.02}$ \\
        \textsc{Bin1K} & 512.66$\pm.00$ & 517.87$\pm.02$ & 517.46$\pm.02$ & 517.08$\pm.02$ & 516.84$\pm.01$ & $1000^*$ & $\bm{509.46}$ \\
        \textsc{Bin2K} & $\bm{1017.88}\pm.00$ & 1030.66$\pm.01$ & 1029.89$\pm.01$ & 1029.11$\pm.02$ &  1028.67$\pm.02$ & $2000^*$ & $1028.67$  \\
    \bottomrule
    \end{tabular}
\end{table*}

\begin{table*}[ht]
    \small
    \caption{PPO results on the Bin Packing benchmark. Lower is better.}
    \label{tab:binpacking-full-ppo}
    \centering
    \begin{tabular}{l | r | rrrr || rr}
    \toprule
        & \multicolumn{1}{| c |}{Greedy} & \multicolumn{4}{| c ||}{Sampled} & OR-Tools & FFD \\
        & $\times 1$ & $\times 1$ & $\times 2$ & $\times 5$ & $\times 10$ & \\
    \midrule
        \textsc{Bin50}  & $27.62\pm.00$ & $27.95\pm.01$ & $27.71\pm.01$ & $27.45\pm.01$ & $27.32\pm.01$ & $\bm{26.71}$ & 27.10 \\
        \textsc{Bin100} & $53.80 \pm .00$ & $54.88 \pm .02$ & $54.27 \pm .02$ & $53.75 \pm .01$ & $53.53 \pm .01$ & $53.91$ & $\bm{52.91}$ \\
        \textsc{Bin200} & $105.63\pm.00$ & $108.51\pm.01$ & $107.20\pm.01$ & $106.21\pm.01$ & $105.86\pm.01$ & $109.19$ & $\bm{104.25}$\\
        \textsc{Bin500} & $259.08\pm.00$ & $268.42\pm.02$ & $264.79\pm.01$ & $262.66\pm.02$ & $261.98\pm.01$ & $267.63$ & $\bm{257.02}$ \\
        \textsc{Bin1K} & $512.66\pm.00$ & $533.97\pm.04$ & $526.23\pm.02$ & $522.30\pm.03$ & $521.22\pm.02$ & $1000^*$ & $\bm{509.46}$ \\
        \textsc{Bin2K} & $\bm{1017.88}\pm.00$ & $1064.74\pm.11$ & $1048.80\pm.06$ & $1041.02\pm.01$ & $1039.09\pm.04$ & $2000^*$ & $1028.67$   \\
    \bottomrule
    \end{tabular}
\end{table*}

\subsection{Travelling Salesperson Problem (TSP)}

\paragraph{Data} We generate random instances for 2D Euclidean TSP by sampling coordinates uniformly in a unit square, as done in previous research \citep{kool2018attention,chen2019learning,costa2020}. We assume complete graphs (fully-connected TSP), which means every pair of cities is connected by a valid route (an edge).

\paragraph{Initial Solution} We start with a random tour, which is simply a random permutation of the city indices. This is likely to be a poor initial solution, as it ignores any information about the problem, namely the coordinates of each city. Nevertheless, Neural SA achieves competitive results in spite of this, and it is reasonable to expect an improvement in its performance (at least in running time) when using better initialisation methods, like in LKH-3 \citep{helsgaun2000effective} for instance.

\paragraph{Training} We train \emph{only on} \textsc{TSP20} with very short rollouts of length $K=40$. Just like in the other problems we consider, we train using 256 random problems generated on the fly for each epoch. We also maintain the same initial temperature and cooling schedule with $T_0=1$ and $\alpha = \left(T_K / T_0 \right)^{\frac{1}{K}}$, but use lower final temperatures for the TSP. We set $T_K=1\mathrm{e}{-2}$ for PPO and $T_K=1\mathrm{e}{-4}$ for ES, which we empirically found to work best with the training dynamics of each of these methods. We also use different number of epochs for each training method, 1000 for PPO and 10 000 for ES, as the latter has slower convergence.

\paragraph{Testing} We evaluate Neural SA on TSP20, TSP50 and TSP100 using the 10K problem instances made available in \cite{kool2018attention}. This allows us to directly compare our methods to previous research on the TSP. We also consider larger problem sizes, namely TSP200 and TSP500 to showcase the scalability of Neural SA. For each of these, we randomly generate 1000 instances by uniformly sampling coordinates in a 2D unit square. For each problem size $N$, we consider rollouts of length $K=N^2$, $K=2N^2$, $K=5N^2$ and $K=10N^2$. That is different from the other CO problems we study since the complexity in the TSP is related to the number of edges $N^2$ rather than the number of cities $N$. We also compare sampled and greedy variants of Neural SA. The former naturally samples actions from the proposal distribution while the latter always selects the most likely action.

We compare Neural SA against standard solvers LKH-3 \citep{helsgaun2000effective} and Concorde \citep{applegate2006traveling}, which we have run ourselves. We also compare against the self-reported results of other Deep Learning models that have targeted TSP and relied on the test data provided by \cite{kool2018attention}: GCN \citep{joshi2019learning}, GAT \citep{kool2018attention}, GAT-T \citep{wu2019learning}, and the works of \cite{costa2020} and \cite{fu2021generalize}. 

Note that \cite{fu2021generalize} also provide results for TSP200 and TSP500, but given that we do not know the exact test instances they used, it is hard to make a direct comparison to our results, especially regarding running times; they use a dataset of 128 instances, while we use 1000. For that reason, we omitted these results from Table~\ref{tab:tsp} in the main text, but for the sake of completeness, presented them in Table~\ref{tab:TSP-full}.

\paragraph{Generalisation} We always train Neural SA only on the smallest of problem sizes we consider. In Table~\ref{tab:generalisation-results}, we compare Neural SA with other models in the literature that have been evaluated the same way: trained on TSP20 only and tested on TSP20, 50 and 100. While not outperforming the model by Fu et al.~\yrcite{fu2021generalize}, Neural SA, especially with PPO, does generalise better than previous end-to-end methods \cite{kool2018attention}.

\begin{table}[!htb]
    \small
    \caption{Optimality gap for models trained on \textsc{TSP20} and evaluated on the test instances provided by Kool et al.~\yrcite{kool2018attention} for \textsc{TSP20/50/100}; *Values taken from respective papers.}
    \label{tab:generalisation-results}
    \centering
    \begin{tabular}{l c c c }
    \toprule
        & \textsc{TSP20} & \textsc{TSP50} & \textsc{TSP100} \\
    \midrule
        Kool et al.~\yrcite{kool2018attention}* &	$0.34\%$ & $\sim 5.0\%$ & $> 14.0\%$ \\
        Fu et al.~\yrcite{fu2021generalize}* &	$0.00\%$ & $0.01\%$	& $0.04\%$ \\
    \midrule
        SA &	$1.17\%$ & $4.34\%$ & $7.43\%$ \\
        Neural SA (PPO) & $0.42\%$ & $1.16\%$ & $1.85\%$ \\
        Neural SA (ES) & $0.10\%$ & $2.32\%$ & $5.50\%$ \\
    \bottomrule
    \end{tabular}
\end{table}

\begin{table*}[ht]
    \small
    \caption{ES results on the TSP benchmark. Lower is better}
    \label{tab:tsp-full-es}
    \centering
    \begin{tabular}{l | r | rrrr || r r}
    \toprule
        & \multicolumn{1}{| c |}{Greedy} & \multicolumn{4}{| c ||}{Sampled} & LKH-3 & Concorde \\
        & $\times 1$ & $\times 1$ & $\times 2$ & $\times 5$ & $\times 10$ & \\
    \midrule
        \textsc{TSP20}    & $3.868 \pm .000$ & $3.868\pm.001$ & $3.854\pm.000$ & $3.844\pm.000$ & $3.840\pm.000$ & \textbf{3.836} & \textbf{3.836} \\
        \textsc{TSP50}    & $6.020 \pm .002$ & $6.022\pm.002$ & $5.947\pm.001$ & $5.871\pm.000$ & $5.828\pm.001$ & \textbf{5.696} & \textbf{5.696}\\
        \textsc{TSP100}   & $8.659 \pm .003$ & $8.660\pm.002$ & $8.477\pm.001$ & $8.298\pm.002$ & $8.191\pm.002$ &  \textbf{7.764} & \textbf{7.764}\\
    \bottomrule
    \end{tabular}
\end{table*}

\begin{table*}[ht]
    \small
    \caption{PPO results on the TSP benchmark. Lower is better}
    \label{tab:tsp-full-ppo}
    \centering
    \begin{tabular}{l | r | rrrr || r r}
    \toprule
        & \multicolumn{1}{| c |}{Greedy} & \multicolumn{4}{| c ||}{Sampled} & LKH-3 & Concorde \\
        & $\times 1$ & $\times 1$ & $\times 2$ & $\times 5$ & $\times 10$ & \\
    \midrule
        \textsc{TSP20}     & $3.864\pm.000$ & $3.865\pm.000$ & $3.850\pm.000$ & $3.841\pm.000$ & $3.838\pm.000$ & \textbf{3.836} & \textbf{3.836} \\
        \textsc{TSP50}    & $5.828\pm.001$ & $5.828\pm.000$ & $5.786\pm.000$ & $5.752\pm.001$ & $5.734\pm.001$ & \textbf{5.696} & \textbf{5.696}\\
        \textsc{TSP100}    & $8.074\pm.001$ & $8.073\pm.001$ & $7.986\pm.001$ & $7.912\pm.001$ & $7.874\pm.000$ &  \textbf{7.764} & \textbf{7.764}\\
        \textsc{TSP200}    & $11.41\pm.00$ & $11.41\pm.00$ & $11.23\pm.00$ & $11.09\pm.00$ & $11.00\pm.00$ &  \textbf{10.70} & \textbf{10.70}\\
        \textsc{TSP500}    & $18.44\pm.002$ & $18.43\pm.001$ & $18.07\pm.003$ & $17.79\pm.006$ & $17.64\pm.003$ &  \textbf{16.54} & \textbf{16.54}\\
    \bottomrule
    \end{tabular}
\end{table*}

\begin{figure*}[ht]
    \centering
    \includegraphics[width=\textwidth]{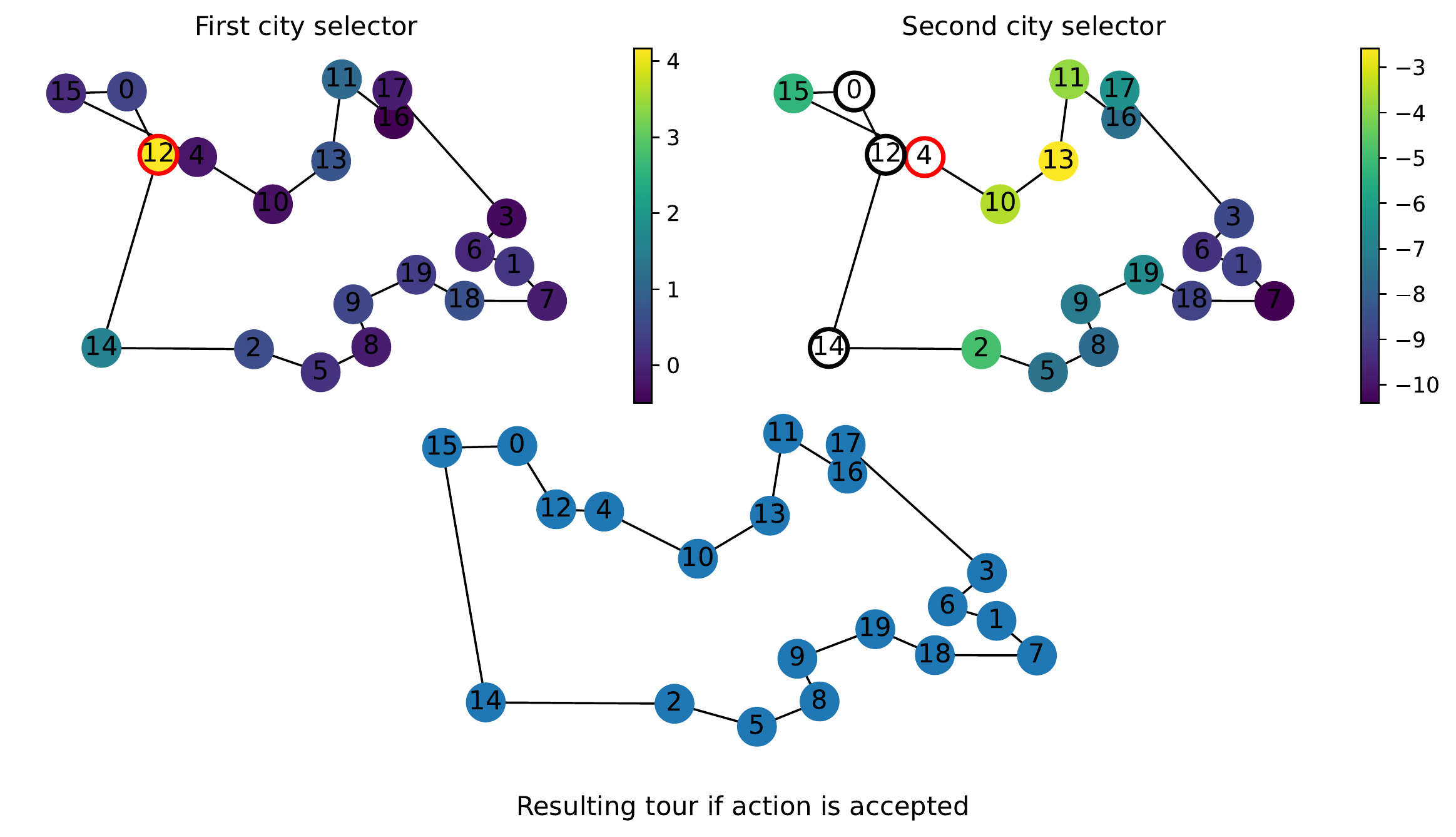}
    \caption{Policy for the Travelling Salesperson Problem. At each step, an action consists of selecting a pair of cities $(i, j)$, one after the other. The figure depicts a TSP problem layed out in the 2D plane, with the learnt proposal distribution over the first city $i$ in the left, and in the right, the distribution over the second city $j$, given $i=12$. We mask out and exclude the neighbours of $i$ ($0$ and $14$) as candidates for $j$ because selecting those would lead to no changes in the tour. It is clear the model has a strong preference towards a few cities, but otherwise the probability mass is spread almost uniformly among the other nodes. However, once $i$ is fixed, Neural SA strongly favours nodes $j$ that are close to $i$. That is a desirable behaviour and even features in popular algorithms like LKH-3 \citep{helsgaun2000effective}. That is because a 2-opt move $(i,j)$ actually adds edge $(i, j)$ to the tour, so leaning towards pairs of cities that are close to each other is more likely to lead to shorter tours.}
    \label{fig:tsp-logits}
\end{figure*}

\vfill

\begin{landscape}
\begin{table}[b]
    \caption{Comparison of different TSP solvers on the 10K instances for TSP20/50/100 provided in \cite{kool2018attention}, and 1K random instances for TSP200/500. We report the average solution cost, optimality gap and running time (to solve all instances) for each problem size. We split competing neural methods in two groups: construction heuristics \cite{kool2018attention,joshi2019efficient} and improvement heuristics like Neural SA \cite{wu2019learning,costa2020,fu2021generalize}.
    *Values as reported in the corresponding paper. $^\dagger$ Different test data.}
    \label{tab:TSP-full}
    \centering
    \resizebox{\columnwidth}{!}{
    \begin{tabular}{l| r r r | r r r | r r r | r r r | r r r }
    \toprule
        &  \multicolumn{3}{| c |}{TSP20} & \multicolumn{ 3 }{| c |}{ TSP50 } & \multicolumn{ 3 }{| c }{TSP100} & \multicolumn{ 3 }{| c |}{ TSP200 } & \multicolumn{ 3 }{| c }{ TSP500 }\\
        & Cost & Gap & Time & Cost & Gap & Time & Cost & Gap & Time & Cost & Gap & Time & Cost & Gap & Time \\
    \midrule
        \textsc{Concorde} \citep{applegate2006traveling} & 3.836 & 0.00\% & 48s & 5.696 & 0.00\% & 2m & 7.764 & 0.00\% & 7m & 10.70 & 0.00\% & 38m & 16.54 & 0.00\% & 7h58m \\
        \textsc{LKH-3} \citep{helsgaun2000effective}    & 3.836 & 0.00\% & 1m & 5.696 & 0.00\% & 14m & 7.764 & 0.00\% & 1h & 10.70 & 0.00\% & 21m & 16.54 & 0.00\% & 1h15m \\
        \textsc{OR-Tools} \citep{ortools} & 3.86 & 0.85\% & 1m & 5.85 & 2.87\% & 5m & 8.06 & 3.86\% & 23m & - & - & - & - & - & - \\
    \midrule
        \textsc{SA}  & 3.881 & 1.17\% & 10s & 5.943 & 4.34\% & 37s & 8.343 & 7.45\% & 3m & 11.98 & 11.87\% & 9m & 20.22 & 22.25\% & 56m \\
        \textsc{Neural SA PPO}    & 3.837 & 0.02\% & 17s & 5.727 & 0.54\% & 1m & 7.856 & 1.18\% & 9m & 10.96 & 2.50\% & 15m & 17.64 & 6.65\% & 2h16m  \\
        \textsc{Neural SA ES}    & 3.840 & 0.10\% & 10s & 5.828 & 2.32\% & 1m & 8.191 & 5.50\% & 9m & 11.74 & 9.72\% & 15m & 20.27 & 22.55\% & 2h16m  \\
    \midrule
        \textsc{GCN} Greedy \citep{joshi2019efficient}* & 3.86 & 0.60\% & 6s & 5.87 & 3.10\% & 55s & 8.41 & 8.38\% & 6m & - & - & - & - & - & -  \\
        \textsc{GCN} Beam Search \citep{joshi2019efficient}* & 3.84 & 0.01\% & 12m & 5.70 & 0.01\% & 18m & 7.87 & 1.39\% & 40m & - & - & - & - & - & -  \\
        \textsc{GAT} Greedy \citep{kool2018attention}* & 3.85 & 0.34\% & 0s & 5.80 & 1.76\% & 2s & 8.12 & 4.53\% & 6s &  - & - & - & - & - & -  \\
        \textsc{GAT} Sampling \citep{kool2018attention}* & 3.84 & 0.08\% & 5 m & 5.73 & 0.52\% & 24m & 7.94 & 2.26\% & 1 h & - & - & - & - & - & -  \\
    \midrule
        \textsc{GAT-T} \{1000\} \citep{wu2019learning}* & 3.84 & 0.03\% & 12m & 5.75 & 0.83\% & 16m & 8.01 & 3.24\% & 25m & - & - & - & - & - & -  \\
        \textsc{GAT-T} \{3000\} \citep{wu2019learning}* & 3.84 & 0.00\% & 39m & 5.72 & 0.34\% & 45 m & 7.91 & 1.85\% & 1 h & - & - & -  & - & - & - \\
        \textsc{GAT-T} \{5000\} \citep{wu2019learning}* & 3.84 & 0.00\% & 1 h & 5.71 & 0.20\% & 1 h & 7.87 & 1.42\% & 2 h & - & - & -  & - & - & -  \\
        \citet{costa2020} \{500\}* & 3.84 & 0.01\% & 5m & 5.72 & 0.36\% & 7m & 7.91 & 1.84\% & 10m & - & - & -  & - & - & -  \\
        \citet{costa2020} \{1000\}* & 3.84 & 0.00\% & 10m & 5.71 & 0.21\% & 13m & 7.86 & 1.26\% & 21 m & - & - & - & - & - & - \\
        \citet{costa2020} \{2000\}* & 3.84 & 0.00\% & 15m & 5.70 & 0.12\% & 29m & 7.83 & 0.87\% & 41m & - & - & - & - & - & - \\
        Att-GCRN+MCTS \citep{fu2021generalize}* & 3.84 & 0.00\% & 2m & 5.69 & 0.01\% & 9m & 7.76 & 0.03\% & 15m & 10.81$^\dagger$ & 0.88\%$^\dagger$ & 3m$^\dagger$ & 16.96$^\dagger$ & 2.96\%$^\dagger$ & 6m$^\dagger$ \\
        
    \bottomrule
    \end{tabular}
    }
\end{table}
\end{landscape}

\end{document}

%% file: math_commands.tex
\usepackage{amsmath,amsfonts,bm}

\def\eqref#1{equation~\ref{#1}}

\def\1{\bm{1}}

\def\rva{{\mathbf{a}}}

\def\rvc{{\mathbf{c}}}

\def\rvs{{\mathbf{s}}}

\def\rvx{{\mathbf{x}}}

\def\rvz{{\mathbf{z}}}

\DeclareMathAlphabet{\mathsfit}{\encodingdefault}{\sfdefault}{m}{sl}
\SetMathAlphabet{\mathsfit}{bold}{\encodingdefault}{\sfdefault}{bx}{n}

\def\gA{{\mathcal{A}}}

\def\gM{{\mathcal{M}}}
\def\gN{{\mathcal{N}}}
\def\gO{{\mathcal{O}}}

\def\gS{{\mathcal{S}}}

\def\gU{{\mathcal{U}}}

\def\gX{{\mathcal{X}}}

\def\sP{{\mathbb{P}}}

\def\sR{{\mathbb{R}}}

\newcommand{\E}{\mathbb{E}}

\DeclareMathOperator*{\argmin}{arg\,min}

%% file: icml2022_conference.bbl
\begin{thebibliography}{66}
\providecommand{\natexlab}[1]{#1}
\providecommand{\url}[1]{\texttt{#1}}
\expandafter\ifx\csname urlstyle\endcsname\relax
  \providecommand{\doi}[1]{doi: #1}\else
  \providecommand{\doi}{doi: \begingroup \urlstyle{rm}\Url}\fi

\bibitem[Albergo et~al.(2019)Albergo, Kanwar, and Shanahan]{albergo2019flow}
Albergo, M., Kanwar, G., and Shanahan, P.
\newblock Flow-based generative models for markov chain monte carlo in lattice
  field theory.
\newblock \emph{Physical Review D}, 100\penalty0 (3):\penalty0 034515, 2019.

\bibitem[Andrychowicz et~al.(2016)Andrychowicz, Denil, Gomez, Hoffman, Pfau,
  Schaul, Shillingford, and De~Freitas]{andrychowicz2016learning}
Andrychowicz, M., Denil, M., Gomez, S., Hoffman, M.~W., Pfau, D., Schaul, T.,
  Shillingford, B., and De~Freitas, N.
\newblock Learning to learn by gradient descent by gradient descent.
\newblock In \emph{Advances in neural information processing systems}, pp.\
  3981--3989, 2016.

\bibitem[Applegate et~al.(2006)Applegate, Bixby, Chv{\'a}tal, and
  Cook]{applegate2006traveling}
Applegate, D.~L., Bixby, R.~E., Chv{\'a}tal, V., and Cook, W.~J.
\newblock \emph{The Traveling Salesman Problem: A Computational Study}.
\newblock Princeton University Press, 2006.

\bibitem[Bello et~al.(2016)Bello, Pham, Le, Norouzi, and Bengio]{bello16}
Bello, I., Pham, H., Le, Q.~V., Norouzi, M., and Bengio, S.
\newblock Neural combinatorial optimization with reinforcement learning.
\newblock \emph{CoRR}, abs/1611.09940, 2016.
\newblock URL \url{http://arxiv.org/abs/1611.09940}.

\bibitem[Beloborodov et~al.(2020)Beloborodov, Ulanov, Foerster, Whiteson, and
  Lvovsky]{beloborodov2020reinforcement}
Beloborodov, D., Ulanov, A.~E., Foerster, J.~N., Whiteson, S., and Lvovsky, A.
\newblock Reinforcement learning enhanced quantum-inspired algorithm for
  combinatorial optimization.
\newblock \emph{Machine Learning: Science and Technology}, 2\penalty0
  (2):\penalty0 025009, 2020.

\bibitem[Bengio et~al.(2018)Bengio, Lodi, and Prouvost]{bengio2018}
Bengio, Y., Lodi, A., and Prouvost, A.
\newblock Machine learning for combinatorial optimization: a methodological
  tour d'horizon.
\newblock \emph{CoRR}, abs/1811.06128, 2018.
\newblock URL \url{http://arxiv.org/abs/1811.06128}.

\bibitem[Berthold(2013)]{bertold2013}
Berthold, T.
\newblock Measuring the impact of primal heuristics.
\newblock \emph{Operations Research Letters}, 41\penalty0 (6):\penalty0
  611--614, 2013.
\newblock ISSN 0167-6377.
\newblock \doi{https://doi.org/10.1016/j.orl.2013.08.007}.
\newblock URL
  \url{https://www.sciencedirect.com/science/article/pii/S0167637713001181}.

\bibitem[Blum et~al.(2020)Blum, Dan, and Seddighin]{blum2020learning}
Blum, A., Dan, C., and Seddighin, S.
\newblock Learning complexity of simulated annealing, 2020.

\bibitem[Bonami et~al.(2018)Bonami, Lodi, and Zarpellon]{bonami2018}
Bonami, P., Lodi, A., and Zarpellon, G.
\newblock Learning a classification of mixed-integer quadratic programming
  problems.
\newblock In van Hoeve, W.-J. (ed.), \emph{Integration of Constraint
  Programming, Artificial Intelligence, and Operations Research}, pp.\
  595--604, Cham, 2018. Springer International Publishing.
\newblock ISBN 978-3-319-93031-2.

\bibitem[Bresson \& Laurent(2021)Bresson and Laurent]{bresson2021transformer}
Bresson, X. and Laurent, T.
\newblock The transformer network for the traveling salesman problem.
\newblock \emph{arXiv preprint arXiv:2103.03012}, 2021.

\bibitem[Cai et~al.(2019)Cai, Hang, Mirhoseini, Tucker, Wang, and
  Wei]{cai2019reinforcement}
Cai, Q., Hang, W., Mirhoseini, A., Tucker, G., Wang, J., and Wei, W.
\newblock Reinforcement learning driven heuristic optimization, 2019.

\bibitem[Chen \& Tian(2019)Chen and Tian]{chen2019learning}
Chen, X. and Tian, Y.
\newblock Learning to perform local rewriting for combinatorial optimization.
\newblock \emph{Advances in Neural Information Processing Systems},
  32:\penalty0 6281--6292, 2019.

\bibitem[Cicirello(2007)]{cicirello2007design}
Cicirello, V.~A.
\newblock On the design of an adaptive simulated annealing algorithm.
\newblock In \emph{Proceedings of the international conference on principles
  and practice of constraint programming first workshop on autonomous search},
  2007.

\bibitem[Croes(1958)]{croes1958method}
Croes, G.~A.
\newblock A method for solving traveling-salesman problems.
\newblock \emph{Operations Research}, 6:\penalty0 791--812, 1958.

\bibitem[da~Costa et~al.(2020)da~Costa, Rhuggenaath, Zhang, and
  Akcay]{costa2020}
da~Costa, P.~R., Rhuggenaath, J., Zhang, Y., and Akcay, A.
\newblock Learning 2-opt heuristics for the traveling salesman problem via deep
  reinforcement learning.
\newblock \emph{CoRR}, abs/2004.01608, 2020.
\newblock URL \url{https://arxiv.org/abs/2004.01608}.

\bibitem[Dai et~al.(2017)Dai, Khalil, Zhang, Dilkina, and Song]{dai2017}
Dai, H., Khalil, E.~B., Zhang, Y., Dilkina, B., and Song, L.
\newblock Learning combinatorial optimization algorithms over graphs.
\newblock \emph{CoRR}, abs/1704.01665, 2017.
\newblock URL \url{http://arxiv.org/abs/1704.01665}.

\bibitem[de~Haan et~al.(2021)de~Haan, Rainone, Cheng, and
  Bondesan]{dehaan2021scaling}
de~Haan, P., Rainone, C., Cheng, M. C.~N., and Bondesan, R.
\newblock Scaling up machine learning for quantum field theory with equivariant
  continuous flows, 2021.

\bibitem[Emami \& Ranka(2018)Emami and Ranka]{emami2018}
Emami, P. and Ranka, S.
\newblock Learning permutations with sinkhorn policy gradient.
\newblock \emph{CoRR}, abs/1805.07010, 2018.
\newblock URL \url{http://arxiv.org/abs/1805.07010}.

\bibitem[Finn et~al.(2017)Finn, Abbeel, and Levine]{finn2017model}
Finn, C., Abbeel, P., and Levine, S.
\newblock Model-agnostic meta-learning for fast adaptation of deep networks.
\newblock In \emph{International Conference on Machine Learning}, pp.\
  1126--1135. PMLR, 2017.

\bibitem[Franzin \& St{\"u}tzle(2019)Franzin and
  St{\"u}tzle]{franzin2019revisiting}
Franzin, A. and St{\"u}tzle, T.
\newblock Revisiting simulated annealing: A component-based analysis.
\newblock \emph{Computers \& operations research}, 104:\penalty0 191--206,
  2019.

\bibitem[Fu et~al.(2021)Fu, Qiu, and Zha]{fu2021generalize}
Fu, Z.-H., Qiu, K.-B., and Zha, H.
\newblock Generalize a small pre-trained model to arbitrarily large tsp
  instances.
\newblock \emph{Proceedings of the AAAI Conference on Artificial Intelligence},
  35\penalty0 (8):\penalty0 7474--7482, May 2021.
\newblock URL \url{https://ojs.aaai.org/index.php/AAAI/article/view/16916}.

\bibitem[Gamrath et~al.(2020{\natexlab{a}})Gamrath, Anderson, Bestuzheva, Chen,
  Eifler, Gasse, Gemander, Gleixner, Gottwald, Halbig, Hendel, Hojny, Koch,
  Le~Bodic, Maher, Matter, Miltenberger, M{\"u}hmer, M{\"u}ller, Pfetsch,
  Schl{\"o}sser, Serrano, Shinano, Tawfik, Vigerske, Wegscheider, Weninger, and
  Witzig]{GamrathEtal2020OO}
Gamrath, G., Anderson, D., Bestuzheva, K., Chen, W.-K., Eifler, L., Gasse, M.,
  Gemander, P., Gleixner, A., Gottwald, L., Halbig, K., Hendel, G., Hojny, C.,
  Koch, T., Le~Bodic, P., Maher, S.~J., Matter, F., Miltenberger, M.,
  M{\"u}hmer, E., M{\"u}ller, B., Pfetsch, M.~E., Schl{\"o}sser, F., Serrano,
  F., Shinano, Y., Tawfik, C., Vigerske, S., Wegscheider, F., Weninger, D., and
  Witzig, J.
\newblock {The SCIP Optimization Suite 7.0}.
\newblock Technical report, Optimization Online, March 2020{\natexlab{a}}.
\newblock URL
  \url{http://www.optimization-online.org/DB_HTML/2020/03/7705.html}.

\bibitem[Gamrath et~al.(2020{\natexlab{b}})Gamrath, Anderson, Bestuzheva, Chen,
  Eifler, Gasse, Gemander, Gleixner, Gottwald, Halbig, Hendel, Hojny, Koch,
  Le~Bodic, Maher, Matter, Miltenberger, M{\"u}hmer, M{\"u}ller, Pfetsch,
  Schl{\"o}sser, Serrano, Shinano, Tawfik, Vigerske, Wegscheider, Weninger, and
  Witzig]{GamrathEtal2020ZR}
Gamrath, G., Anderson, D., Bestuzheva, K., Chen, W.-K., Eifler, L., Gasse, M.,
  Gemander, P., Gleixner, A., Gottwald, L., Halbig, K., Hendel, G., Hojny, C.,
  Koch, T., Le~Bodic, P., Maher, S.~J., Matter, F., Miltenberger, M.,
  M{\"u}hmer, E., M{\"u}ller, B., Pfetsch, M.~E., Schl{\"o}sser, F., Serrano,
  F., Shinano, Y., Tawfik, C., Vigerske, S., Wegscheider, F., Weninger, D., and
  Witzig, J.
\newblock {The SCIP Optimization Suite 7.0}.
\newblock ZIB-Report 20-10, Zuse Institute Berlin, March 2020{\natexlab{b}}.
\newblock URL \url{http://nbn-resolving.de/urn:nbn:de:0297-zib-78023}.

\bibitem[Gasse et~al.(2019)Gasse, Chetelat, Ferroni, Charlin, and
  Lodi]{gasse2019exact}
Gasse, M., Chetelat, D., Ferroni, N., Charlin, L., and Lodi, A.
\newblock Exact combinatorial optimization with graph convolutional neural
  networks.
\newblock In Wallach, H., Larochelle, H., Beygelzimer, A., d\textquotesingle
  Alch\'{e}-Buc, F., Fox, E., and Garnett, R. (eds.), \emph{Advances in Neural
  Information Processing Systems}, volume~32. Curran Associates, Inc., 2019.
\newblock URL
  \url{https://proceedings.neurips.cc/paper/2019/file/d14c2267d848abeb81fd590f371d39bd-Paper.pdf}.

\bibitem[Geman \& Geman(1984)Geman and Geman]{geman1984stochastic}
Geman, S. and Geman, D.
\newblock Stochastic relaxation, gibbs distributions, and the bayesian
  restoration of images.
\newblock \emph{IEEE Transactions on Pattern Analysis and Machine
  Intelligence}, PAMI-6\penalty0 (6):\penalty0 721--741, 1984.
\newblock \doi{10.1109/TPAMI.1984.4767596}.

\bibitem[Gupta et~al.(2020)Gupta, Gasse, Khalil, Kumar, Lodi, and
  Bengio]{gupta2020}
Gupta, P., Gasse, M., Khalil, E.~B., Kumar, M.~P., Lodi, A., and Bengio, Y.
\newblock Hybrid models for learning to branch.
\newblock \emph{CoRR}, abs/2006.15212, 2020.
\newblock URL \url{https://arxiv.org/abs/2006.15212}.

\bibitem[Hastings(1970)]{hastings1970}
Hastings, W.~K.
\newblock {Monte Carlo sampling methods using Markov chains and their
  applications}.
\newblock \emph{Biometrika}, 57\penalty0 (1):\penalty0 97--109, 04 1970.
\newblock ISSN 0006-3444.
\newblock \doi{10.1093/biomet/57.1.97}.
\newblock URL \url{https://doi.org/10.1093/biomet/57.1.97}.

\bibitem[Helsgaun(2000)]{helsgaun2000effective}
Helsgaun, K.
\newblock An effective implementation of the lin--kernighan traveling salesman
  heuristic.
\newblock \emph{European Journal of Operational Research}, 126\penalty0
  (1):\penalty0 106--130, 2000.

\bibitem[Holland(1992)]{holland1993}
Holland, J.~H.
\newblock \emph{Adaptation in Natural and Artificial Systems: An Introductory
  Analysis with Applications to Biology, Control and Artificial Intelligence}.
\newblock MIT Press, Cambridge, MA, USA, 1992.
\newblock ISBN 0262082136.

\bibitem[Ingber(1996)]{ingber1996adaptive}
Ingber, L.
\newblock Adaptive simulated annealing (asa): lessons learned.
\newblock \emph{Control and Cybernetics}, 25\penalty0 (1), 1996.

\bibitem[Ji et~al.(2021)Ji, Yang, and Liang]{ji2021bilevel}
Ji, K., Yang, J., and Liang, Y.
\newblock Bilevel optimization: Convergence analysis and enhanced design.
\newblock In \emph{International Conference on Machine Learning}, pp.\
  4882--4892. PMLR, 2021.

\bibitem[Johnson(1973)]{johnson1973near}
Johnson, D.~S.
\newblock \emph{Near-optimal bin packing algorithms}.
\newblock PhD thesis, Massachusetts Institute of Technology, 1973.

\bibitem[Joshi et~al.(2019{\natexlab{a}})Joshi, Laurent, and
  Bresson]{joshi2019efficient}
Joshi, C.~K., Laurent, T., and Bresson, X.
\newblock An efficient graph convolutional network technique for the travelling
  salesman problem.
\newblock \emph{arXiv preprint arXiv:1906.01227}, 2019{\natexlab{a}}.

\bibitem[Joshi et~al.(2019{\natexlab{b}})Joshi, Laurent, and
  Bresson]{joshi2019learning}
Joshi, C.~K., Laurent, T., and Bresson, X.
\newblock On learning paradigms for the travelling salesman problem.
\newblock \emph{arXiv preprint arXiv:1910.07210}, 2019{\natexlab{b}}.

\bibitem[Khairy et~al.(2020)Khairy, Shaydulin, Cincio, Alexeev, and
  Balaprakash]{Khairy_2020}
Khairy, S., Shaydulin, R., Cincio, L., Alexeev, Y., and Balaprakash, P.
\newblock Learning to optimize variational quantum circuits to solve
  combinatorial problems.
\newblock \emph{Proceedings of the AAAI Conference on Artificial Intelligence},
  34\penalty0 (03):\penalty0 2367–2375, Apr 2020.
\newblock ISSN 2159-5399.
\newblock \doi{10.1609/aaai.v34i03.5616}.
\newblock URL \url{http://dx.doi.org/10.1609/aaai.v34i03.5616}.

\bibitem[Kingma \& Ba(2015)Kingma and Ba]{kingma2015adam}
Kingma, D.~P. and Ba, J.
\newblock Adam: {A} method for stochastic optimization.
\newblock In Bengio, Y. and LeCun, Y. (eds.), \emph{3rd International
  Conference on Learning Representations, {ICLR} 2015, San Diego, CA, USA, May
  7-9, 2015, Conference Track Proceedings}, 2015.
\newblock URL \url{http://arxiv.org/abs/1412.6980}.

\bibitem[Kirkpatrick et~al.(1987)Kirkpatrick, Gelatt~Jr, and
  Vecchi]{kirkpatrick1987optimization}
Kirkpatrick, S., Gelatt~Jr, C.~D., and Vecchi, M.~P.
\newblock Optimization by simulated annealing.
\newblock In \emph{Readings in Computer Vision}, pp.\  606--615. Elsevier,
  1987.

\bibitem[Kool et~al.(2018)Kool, van Hoof, and Welling]{kool2018attention}
Kool, W., van Hoof, H., and Welling, M.
\newblock Attention, learn to solve routing problems!
\newblock In \emph{International Conference on Learning Representations}, 2018.

\bibitem[Kool et~al.(2021)Kool, van Hoof, Gromicho, and Welling]{kool2021deep}
Kool, W., van Hoof, H., Gromicho, J., and Welling, M.
\newblock Deep policy dynamic programming for vehicle routing problems.
\newblock \emph{arXiv preprint arXiv:2102.11756}, 2021.

\bibitem[Kruber et~al.(2017)Kruber, Lübbecke, and Parmentier]{kruber2017}
Kruber, M., Lübbecke, M., and Parmentier, A.
\newblock Learning when to use a decomposition.
\newblock In \emph{CPAIOR}, pp.\  202--210, 05 2017.
\newblock ISBN 978-3-319-59775-1.
\newblock \doi{10.1007/978-3-319-59776-8_16}.

\bibitem[Lam \& Delosme(1988)Lam and Delosme]{lam1988performance}
Lam, J. and Delosme, J.-M.
\newblock Performance of a new annealing schedule.
\newblock In \emph{Proceedings of the 25th ACM/IEEE Design Automation
  Conference}, pp.\  306--311, 1988.

\bibitem[Likhosherstov et~al.(2021)Likhosherstov, Song, Choromanski, Davis, and
  Weller]{likhosherstov2021debiasing}
Likhosherstov, V., Song, X., Choromanski, K., Davis, J., and Weller, A.
\newblock Debiasing a first-order heuristic for approximate bi-level
  optimization.
\newblock \emph{arXiv preprint arXiv:2106.02487}, 2021.

\bibitem[Maclaurin et~al.(2015)Maclaurin, Duvenaud, and
  Adams]{maclaurin2015gradientbased}
Maclaurin, D., Duvenaud, D., and Adams, R.~P.
\newblock Gradient-based hyperparameter optimization through reversible
  learning, 2015.

\bibitem[Marcos~Alvarez et~al.(2012)Marcos~Alvarez, Maes, and
  Wehenkel]{marcos2012supervised}
Marcos~Alvarez, A., Maes, F., and Wehenkel, L.
\newblock Supervised learning to tune simulated annealing for in silico protein
  structure prediction.
\newblock In \emph{ESANN 2012 proceedings, 20th European Symposium on
  Artificial Neural Networks, Computational Intelligence and Machine Learning},
  pp.\  49--54. Ciaco, 2012.

\bibitem[Metropolis et~al.(1953)Metropolis, Rosenbluth, Rosenbluth, Teller, and
  Teller]{metropolis1953}
Metropolis, N., Rosenbluth, A.~W., Rosenbluth, M.~N., Teller, A.~H., and
  Teller, E.
\newblock Equation of state calculations by fast computing machines.
\newblock \emph{The Journal of Chemical Physics}, 21\penalty0 (6):\penalty0
  1087--1092, 1953.
\newblock \doi{10.1063/1.1699114}.
\newblock URL \url{https://doi.org/10.1063/1.1699114}.

\bibitem[Mills et~al.(2020)Mills, Ronagh, and Tamblyn]{mills2020finding}
Mills, K., Ronagh, P., and Tamblyn, I.
\newblock Finding the ground state of spin hamiltonians with reinforcement
  learning.
\newblock \emph{Nature Machine Intelligence}, 2\penalty0 (9):\penalty0
  509--517, 2020.

\bibitem[No{\'e} et~al.(2019)No{\'e}, Olsson, K{\"o}hler, and
  Wu]{noe2019boltzmann}
No{\'e}, F., Olsson, S., K{\"o}hler, J., and Wu, H.
\newblock Boltzmann generators: Sampling equilibrium states of many-body
  systems with deep learning.
\newblock \emph{Science}, 365\penalty0 (6457), 2019.

\bibitem[Nomer et~al.(2020)Nomer, Alnowibet, Elsayed, and
  Mohamed]{nomer2020neural}
Nomer, H.~A., Alnowibet, K.~A., Elsayed, A., and Mohamed, A.~W.
\newblock Neural knapsack: A neural network based solver for the knapsack
  problem.
\newblock \emph{IEEE Access}, 8:\penalty0 224200--224210, 2020.

\bibitem[Paszke et~al.(2019)Paszke, Gross, Massa, Lerer, Bradbury, Chanan,
  Killeen, Lin, Gimelshein, Antiga, Desmaison, Kopf, Yang, DeVito, Raison,
  Tejani, Chilamkurthy, Steiner, Fang, Bai, and Chintala]{paszke2017automatic}
Paszke, A., Gross, S., Massa, F., Lerer, A., Bradbury, J., Chanan, G., Killeen,
  T., Lin, Z., Gimelshein, N., Antiga, L., Desmaison, A., Kopf, A., Yang, E.,
  DeVito, Z., Raison, M., Tejani, A., Chilamkurthy, S., Steiner, B., Fang, L.,
  Bai, J., and Chintala, S.
\newblock Pytorch: An imperative style, high-performance deep learning library.
\newblock In Wallach, H., Larochelle, H., Beygelzimer, A., d\textquotesingle
  Alch\'{e}-Buc, F., Fox, E., and Garnett, R. (eds.), \emph{Advances in Neural
  Information Processing Systems 32}, pp.\  8024--8035. Curran Associates,
  Inc., 2019.

\bibitem[Pereira \& Fernandes(2004)Pereira and Fernandes]{pereira2004study}
Pereira, A.~I. and Fernandes, E. M. G.~P.
\newblock A study of simulated annealing variants.
\newblock In \emph{Proceedings of XXVIII Congreso de Estadística e
  Investigación Operativa}, 2004.

\bibitem[Perron \& Furnon(2019)Perron and Furnon]{ortools}
Perron, L. and Furnon, V.
\newblock Or-tools, 2019.
\newblock URL \url{https://developers.google.com/optimization/}.

\bibitem[Rere et~al.(2015)Rere, Fanany, and Arymurthy]{rasdi2015simulated}
Rere, L.~R., Fanany, M.~I., and Arymurthy, A.~M.
\newblock Simulated annealing algorithm for deep learning.
\newblock \emph{Procedia Computer Science}, 72:\penalty0 137--144, 2015.
\newblock ISSN 1877-0509.
\newblock \doi{https://doi.org/10.1016/j.procs.2015.12.114}.
\newblock The Third Information Systems International Conference 2015.

\bibitem[Rieck(2021)]{rieck2021basic}
Rieck, B.
\newblock Basic analysis of bin-packing heuristics.
\newblock \emph{arXiv preprint arXiv:2104.12235}, 2021.

\bibitem[Salimans et~al.(2017)Salimans, Ho, Chen, and
  Sutskever]{salimans2017evolution}
Salimans, T., Ho, J., Chen, X., and Sutskever, I.
\newblock Evolution strategies as a scalable alternative to reinforcement
  learning.
\newblock \emph{ArXiv}, abs/1703.03864, 2017.

\bibitem[Salimifard et~al.(2012)Salimifard, Shahbandarzadeh, and
  Raeesi]{salimifard2012green}
Salimifard, K., Shahbandarzadeh, H., and Raeesi, R.
\newblock Green transportation and the role of operations research.
\newblock In \emph{2012 International Conference on Traffic and Transportation
  Engineering (ICTTE 2012)}, pp.\  74--79, 2012.

\bibitem[Schulman et~al.(2016)Schulman, Moritz, Levine, Jordan, and
  Abbeel]{schulman2016high}
Schulman, J., Moritz, P., Levine, S., Jordan, M., and Abbeel, P.
\newblock High-dimensional continuous control using generalized advantage
  estimation.
\newblock In \emph{Proceedings of the International Conference on Learning
  Representations (ICLR)}, 2016.

\bibitem[Schulman et~al.(2017)Schulman, Wolski, Dhariwal, Radford, and
  Klimov]{schulman2017proximal}
Schulman, J., Wolski, F., Dhariwal, P., Radford, A., and Klimov, O.
\newblock Proximal policy optimization algorithms.
\newblock \emph{arXiv preprint arXiv:1707.06347}, 2017.

\bibitem[Swendsen \& Wang(1986)Swendsen and Wang]{swendsenwang1986}
Swendsen, R.~H. and Wang, J.-S.
\newblock Replica monte carlo simulation of spin-glasses.
\newblock \emph{Phys. Rev. Lett.}, 57:\penalty0 2607--2609, Nov 1986.
\newblock \doi{10.1103/PhysRevLett.57.2607}.
\newblock URL \url{https://link.aps.org/doi/10.1103/PhysRevLett.57.2607}.

\bibitem[van Laarhoven \& Aarts(1987)van Laarhoven and Aarts]{van1987simulated}
van Laarhoven, P. and Aarts, E.
\newblock \emph{Simulated Annealing: Theory and Applications}, chapter 3, Thm.
  6.
\newblock Mathematics and Its Applications. Springer Netherlands, 1987.
\newblock ISBN 9789027725134.
\newblock URL \url{https://books.google.co.in/books?id=-IgUab6Dp\_IC}.

\bibitem[Vashisht et~al.(2020)Vashisht, Rampal, Liao, Lu, Shanbhag, Fallon, and
  Kara]{vashisht2020placement}
Vashisht, D., Rampal, H., Liao, H., Lu, Y., Shanbhag, D., Fallon, E., and Kara,
  L.~B.
\newblock Placement in integrated circuits using cyclic reinforcement learning
  and simulated annealing, 2020.

\bibitem[Vicol et~al.(2021)Vicol, Metz, and Sohl-Dickstein]{vicol2021unbiased}
Vicol, P., Metz, L., and Sohl-Dickstein, J.
\newblock Unbiased gradient estimation in unrolled computation graphs with
  persistent evolution strategies.
\newblock In \emph{International Conference on Machine Learning}, pp.\
  10553--10563. PMLR, 2021.

\bibitem[Vinyals et~al.(2017)Vinyals, Fortunato, and
  Jaitly]{vinyals2017pointer}
Vinyals, O., Fortunato, M., and Jaitly, N.
\newblock Pointer networks, 2017.

\bibitem[Wauters et~al.(2020)Wauters, Panizon, Mbeng, and
  Santoro]{Wauters_2020}
Wauters, M.~M., Panizon, E., Mbeng, G.~B., and Santoro, G.~E.
\newblock Reinforcement-learning-assisted quantum optimization.
\newblock \emph{Physical Review Research}, 2\penalty0 (3), Sep 2020.
\newblock ISSN 2643-1564.
\newblock \doi{10.1103/physrevresearch.2.033446}.
\newblock URL \url{http://dx.doi.org/10.1103/PhysRevResearch.2.033446}.

\bibitem[Wu et~al.(2019{\natexlab{a}})Wu, Song, Cao, Zhang, and
  Lim]{wu2019learning}
Wu, Y., Song, W., Cao, Z., Zhang, J., and Lim, A.
\newblock Learning improvement heuristics for solving the travelling salesman
  problem.
\newblock \emph{CoRR}, abs/1912.05784, 2019{\natexlab{a}}.
\newblock URL \url{http://arxiv.org/abs/1912.05784}.

\bibitem[Wu et~al.(2019{\natexlab{b}})Wu, Song, Cao, Zhang, and
  Lim]{wu2021learning}
Wu, Y., Song, W., Cao, Z., Zhang, J., and Lim, A.
\newblock Learning improvement heuristics for solving the travelling salesman
  problem.
\newblock \emph{CoRR}, abs/1912.05784, 2019{\natexlab{b}}.
\newblock URL \url{http://arxiv.org/abs/1912.05784}.

\bibitem[Zaheer et~al.(2017)Zaheer, Kottur, Ravanbakhsh, Poczos, Salakhutdinov,
  and Smola]{zaheer2017deep}
Zaheer, M., Kottur, S., Ravanbakhsh, S., Poczos, B., Salakhutdinov, R.~R., and
  Smola, A.~J.
\newblock Deep sets.
\newblock \emph{Advances in Neural Information Processing Systems}, 30, 2017.

\end{thebibliography}
